\documentclass[runningheads]{llncs}
\usepackage{graphicx}

\usepackage{tikz}
\usepackage{comment}
\usepackage{amsmath,amssymb} 
\usepackage{color}
\usepackage{graphicx}
\usepackage{caption}
\usepackage{subcaption}
\usepackage{multirow}
\usepackage{bbding}
\usepackage{wrapfig}
\usepackage{booktabs}
\usepackage[colorlinks, linkcolor=black, anchorcolor=black, citecolor=black, urlcolor=black]{hyperref}
\usepackage{orcidlink}
\newcommand{\tabincell}[2]{\begin{tabular}{@{}#1@{}}#2\end{tabular}}
\newcommand{\etal}{\emph{et al. }}           

\usepackage[accsupp]{axessibility}  

\begin{document}
\pagestyle{headings}
\mainmatter
\def\ECCVSubNumber{6366}  

\title{Few-Shot Class-Incremental Learning from an Open-Set Perspective} 

\titlerunning{Few-Shot Class-Incremental Learning from an Open-Set Perspective}
%
\author{Can Peng\inst{1}\orcidlink{0000-0003-1673-2460}\and
Kun Zhao\inst{2} \and
Tianren Wang\inst{1} \and
Meng Li\inst{1} \and
Brian C. Lovell\inst{1}
}
\authorrunning{C. Peng et al.}
%
\institute{The University of Queensland, Brisbane, QLD, Australia \and
Sullivan Nicolaides Pathology, Australia \\
\email{can.peng@uq.net.au, kun\_zhao@snp.com.au, tianren.wang@uq.net.au, meng.li6@uq.net.au, lovell@itee.uq.edu.au} 
}
\maketitle

\begin{abstract}
The continual appearance of new objects in the visual world poses considerable challenges for current deep learning methods in real-world deployments.
The challenge of new task learning is often exacerbated by the scarcity of data for the new categories due to rarity or cost.
Here we explore the important task of Few-Shot Class-Incremental Learning (FSCIL) and its extreme data scarcity condition of one-shot.
An ideal FSCIL model needs to perform well on all classes, regardless of their presentation order or paucity of data.
It also needs to be robust to open-set real-world conditions and be easily adapted to the new tasks that always arise in the field.
In this paper, we first reevaluate the current task setting and propose a more comprehensive and practical setting for the FSCIL task.
Then, inspired by the similarity of the goals for FSCIL and modern face recognition systems, we propose our method --- Augmented Angular Loss Incremental Classification or ALICE.
In ALICE, instead of the commonly used cross-entropy loss, we propose to use the angular penalty loss to obtain well-clustered features.
As the obtained features not only need to be compactly clustered but also diverse enough to maintain generalization for future incremental classes, we further discuss how class augmentation, data augmentation, and data balancing affect classification performance.
Experiments on benchmark datasets, including CIFAR100, miniImageNet, and CUB200, demonstrate the improved performance of ALICE over the state-of-the-art FSCIL methods.
Code is available at \href{https://github.com/CanPeng123/FSCIL_ALICE.git}{https://github.com/CanPeng123/FSCIL\_ALICE}.
\end{abstract}

\section{Introduction}
\label{sec: Introduction}
In recent years, the computer vision community has witnessed astonishing performance breakthroughs in many traditional vision tasks.
These breakthroughs are mainly due to the emergence of deep learning models and algorithms, publicly available large data sets for training, and powerful GPU computing devices.
Despite their popularity, current deep learning techniques mostly rely on large-scale supervised data to train accurate models.
A deep neural network (DNN) with tens of thousands of parameters cannot be easily adapted to a new task by training on just a few examples.
In addition, conventional deep learning models lack the capability of preserving previous knowledge while adapting to new tasks.
When a neural network is fine-tuned to learn a new task, its performance on previously trained tasks will significantly deteriorate, a problem known as catastrophic forgetting \cite{goodfellow2013empirical,mccloskey1989catastrophic}.
Exploring the fast learning and memorizing capability of deep learning models is an important step toward improving their practical application ability.

In this paper, we tackle this significant research direction --- Few-Shot Class-Incremental Learning (FSCIL).
FSCIL requires the trained model to not only quickly adapt to continually arriving new tasks, but also to retain the old knowledge about previously learned tasks.
Considering real-life application, an ideal FSCIL model needs to have the following characteristics:
1) The model needs to perform well on all classes equally, no matter what the training presentation sequence is; and
2) the model needs to be robust to extreme data scarcity, such as the one-shot scenario.
However, current SOTA methods mainly use sole class-wise average accuracy to evaluate the model performance which cannot assess whether there is a prediction bias due to class imbalance and data imbalance.
As there are normally more base classes than incremental classes and only limited data is provided for each incremental class, prediction bias towards base classes can easily happen.
In addition, current SOTA methods rarely consider the extreme one-shot setting which can happen in the real world due to incremental data collection and rare data types.
A well-established task setup is a cornerstone for the development of this task since an improper task setup will misguide the method design and lead to methods with limited application.
Thus, before designing our method, we reformulate the setup for the FSCIL task.

Considering the paucity of incremental session data and the absence of old session data, we think the feature extractor trained on the base session should not be limited to extracting discriminative features for the base categories.
The ability of representing new unseen samples from future novel classes is also critical.
On the one hand, we are motivated by the similarity between FSCIL and face recognition tasks.
The face recognition system learns to distinguish and recognize new faces quickly via its deep metric learning framework.
The capability of handling new identities without the need for retraining is a major achievement of modern face recognition methods and is also what the FSCIL task desires.
On the other hand, we are motivated by the intuitive connection between FSCIL and data augmentation.
Data augmentation focuses on improving the generalization of a DNN. 
The capability of extracting diverse features that is transferable across base and incremental classes is important for the FSCIL task.
Hence in this work, we adopt some ideas from both modern face recognition and data augmentation to design our method.

\textbf{The contributions} of this paper are:
\textbf{(1)} We reevaluate the current benchmark task settings of FSCIL and propose additional experimental settings and evaluation metrics to more comprehensively assess the capability of FSCIL methods. 
\textbf{(2)} We solve the FSCIL task from a new perspective of the open-set problem. 
We analyze the angular penalty loss from face recognition and adapt it to FSCIL to improve the discrimination of the model.
\textbf{(3)} We further analyze how data processing, such as class augmentation, data augmentation, and balanced data embedding affect FSCIL performance and aim to improve the generalization of the model.
\textbf{(4)} Significant improvements on three benchmark datasets, CIFAR100, miniImageNet, and CUB200, demonstrate the effectiveness of our method against SOTA methods.

\section{Related Work}
\label{sec: Related Work}
\noindent\textbf{Few-shot Class-incremental Learning.}
The FSCIL task is a newly emerged challenge evolved from class-incremental learning \cite{rebuffi2017icarl,castro2018end,hou2019learning}.
Once established, the research community has spent much effort developing algorithms for this important FSCIL task.
For SOTA FSCIL methods, after base session training, some update the backbone \cite{tao2020few,cheraghian2021semantic,zhao2020mgsvf,dong2021few} and some freeze the backbone \cite{zhang2021few,zhu2021self,cheraghian2021synthesized}.
Backbone updating methods commonly use the knowledge distillation \cite{hinton2015distilling} technique to preserve the old knowledge. 
Knowledge distillation relies on having sufficient data to simulate the input-output function of the old model.
To adapt knowledge distillation to FSCIL, these methods store old exemplars, require a complex updating scheme for each incremental task, or are incapable of extreme data scarcity conditions such as 1-shot.
However, high performance and flexible operation are both important for real-world applications.
Also, storing old exemplars is undesirable due to memory restrictions. 
In addition, the backbone network has a large number of parameters despite there being extremely limited new task data.
The large imbalance between parameters and data causes the backbone updating methods to normally show lower performance than backbone freezing methods under the same experimental setup.

On the contrary, freezing the backbone network is a good choice to well balance not only the real-life application requirements but also the stability and plasticity trade-off.
This backbone freezing strategy decouples the learning of representations and classifiers to avoid overfitting and catastrophic forgetting in the representations. 
Also, the fundamental feature characteristics are similar for many objects, so features learned from the base session can be readopted for recognizing new classes. 
Our method belongs to the backbone freezing type of methods.
Although this decoupling strategy has been explored by Zhang \etal \cite{zhang2021few}, their method focuses on designing a discriminative classifier.
On the contrary, we focus on feature distribution, since this is a cornerstone of robust classification performance.
Last but not the least, a good FSCIL method needs to perform equally well on all the classes no matter whether they are base or incremental classes.
This is a problem for the current backbone freezing type of methods that their good overall accuracy is mainly derived from the base session. 
In this paper, we target on proposing an FSCIL method that takes advantage of decoupling representation and classification via backbone freezing, and at the same time, solves the side effect of prediction bias. 

\noindent\textbf{Deep Metric Learning.}
Deep metric learning is commonly used for face recognition tasks.
Inspired by the relation between normalized weights on the last fully connected layer and class centers, Liu \etal proposed SphereFace \cite{liu2017sphereface} which uses an angular margin penalty to enforce extra intra-class compactness and inter-class discrepancy.
Following SphereFace, CosFace \cite{wang2018cosface} and ArcFace \cite{deng2019arcface} were proposed to reduce the complex loss calculation and make the training procedure more stable.
There are many similarities between face recognition and FSCIL tasks:
1) both tasks are open-set object recognition tasks that need to classify a large amount of continually arriving new objects (classes/face identities); 
2) both tasks are provided with unbalanced data; and
3) both tasks require fast adaptation on new objects as well as maintaining performance on old objects.
Inspired by these similarities, in this paper, we try to solve the FSCIL task from a new perspective of the open-set problem. 
We adopt the idea of angular penalty loss from face recognition to the more general problem of object recognition.

As real-world classification problems typically exhibit class imbalance or long-tailed data distribution, some methods have explored deep metric learning for incremental and long-tail tasks \cite{yu2020semantic,mai2021supervised,kang2019decoupling}.
However, these methods normally assume sufficient data is available which is a different setting from FSCIL.
Most FSCIL methods solve this problem from the perspective of either incremental learning (advanced knowledge distillation) \cite{cheraghian2021semantic,zhao2020mgsvf,dong2021few} or few-shot learning (freezing backbone and evolving prototypes) \cite{zhang2021few,cheraghian2021synthesized,zhu2021self}. 
We follow the proposal of freezing the backbone network to decouple the learning of representations and classifiers. 
However, different from current backbone freezing type of methods that maintain the incremental learning ability by evolving classification prototypes, we focus on improving the transfer capability of the feature extractor.

\section{Problem Formulation}
\label{sec: Problem Formulation}
\begin{figure}[tbp]
	\centering
	\includegraphics[width=12cm,keepaspectratio]{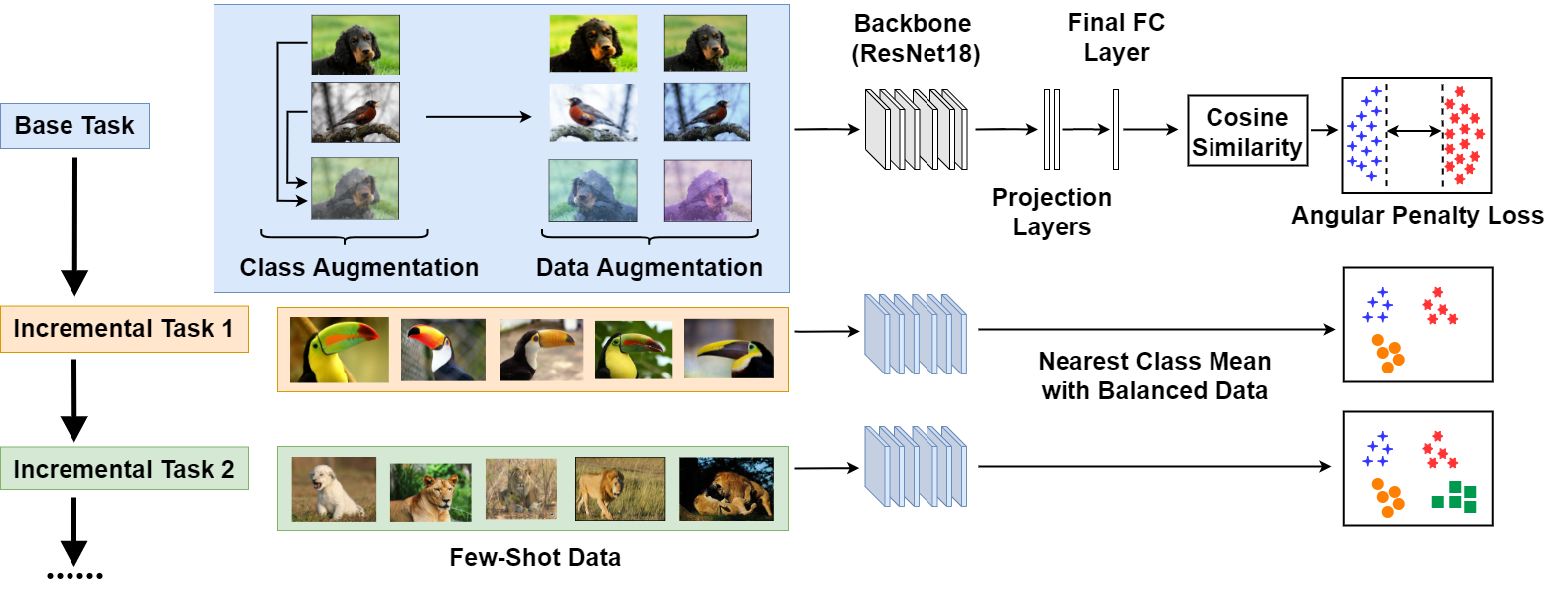}
	\caption{The framework of our proposed method.
		On the one hand, with sufficient base task data available, angular penalty loss, class augmentation, and data augmentation are utilized to obtain a general open-set feature extractor.
		On the other hand, as only limited incremental task data is available, the few-shot new class data and the carefully chosen same number of base class data are utilized to generate the balanced class-wise prototypes.
		Nearest class mean and cosine similarity are adopted to do the final classification.
	}
	\label{fig: framework}	
\end{figure}

FSCIL task comprises a base task with sufficient training data and multiple incremental tasks with limited training data.
During the learning of each new task, only the data for the current task is available and the model is required to learn this new task information whilst retaining old task knowledge.

To be specific, assume an $M$-step FSCIL task.
Let $\{D^0_{train}, D^1_{train}, ..., D^m_{train}\}$ and $\{D^0_{test}, D^1_{test}, ..., D^m_{test}\}$ denotes the training and testing data for sessions $\{0, 1, ..., m\}$, respectively.
For session $i$, it has training data $D^i_{train}$ with the corresponding label space of $C^i$.
Training data from different sessions have no overlapped classes, so when $i \neq j$, $C^i \cap C^j = \varnothing$.
During testing, the model will be evaluated on all seen classes so far, so for session $i$, its testing data $D^i_{test}$ has the corresponding label space of $C^0 \cup C^1 ... \cup C^i$.
In addition, for the base session ($i = 0$), a sufficient amount of training data is provided and for the following incremental sessions ($i > 0 $), only a limited amount of data is provided.

Most papers about FSCIL \cite{cheraghian2021semantic,zhao2020mgsvf,dong2021few,zhang2021few,zhu2021self,cheraghian2021synthesized} follow the task setting proposed by Tao \etal \cite{tao2020few}.
As FSCIL focuses on mimicking real-life situations, we think some aspects of the current benchmark experimental protocol are not sufficient to evaluate the efficiency of an FSCIL method.
Thus, before proposing our method, we propose a more comprehensive and practical setup for the FSCIL task. 

\noindent\textbf{Number of Few-Shot Data.}
Current benchmark experiments are performed with 5-shot, 10-shot, or more data being available for each incremental step. 
The extreme data scarcity condition of 1-shot which can easily happen in the real world due to extremely scarce data type is rarely considered.

\noindent\textbf{Evaluation Metric.}
Current benchmark evaluation metrics mainly use class-wise average accuracy to evaluate the performance of an FSCIL model.
As there are normally more base classes than incremental new classes, using average accuracy cannot indicate if there is a prediction bias between base and incremental classes.
A method cannot be regarded as a good FSCIL method if its good performance is mainly determined by the base class performance. 

\noindent\textbf{Dataset.}
The similarity between base classes and new classes will strongly affect model performance since the high re-usability of base features such as fine-grained datasets will naturally reduce the challenge of catastrophic forgetting.
An optimal FSCIL model needs to not only perform well on high-distributional-match fine-grained datasets but also on low-distributional-match datasets. 

\noindent To sum up, to comprehensively simulate the real-world FSCIL condition and evaluate the robustness of an FSCIL method, we consider both benchmark 5-shot and 1-shot settings. 
Also, for the evaluation metric, we propose to use both average accuracy and harmonic accuracy to evaluate not only the overall performance but also the performance balance between base and incremental classes.
In addition, we perform experiments on both general (CIFAR100 and mini-ImageNet) and fine-grained (CUB200) datasets to remove the possible performance benefit due to high similarity between base and incremental classes.

\section{Methodology}
\label{sec: Methodology}
In this section, we propose the FSCIL method ALICE using angular penalty, class and data augmentation, and data balancing.
First, for the base session, we apply the angular penalty loss to train the feature extractor to obtain compact intra-class clustering and wide inter-class separation.
Class augmentation and data augmentation are also adopted to improve the generalization of the feature extractor.
Then, for the incremental sessions, specifically chosen balanced data are utilized to generate prototypes for each class.
Nearest class mean and cosine similarity are combined to perform the classification.
Figure \ref{fig: framework} demonstrates the framework of our method.

\subsection{Angular Penalty}
\begin{figure}[tbp]
	\centering
	\includegraphics[width=12cm,keepaspectratio]{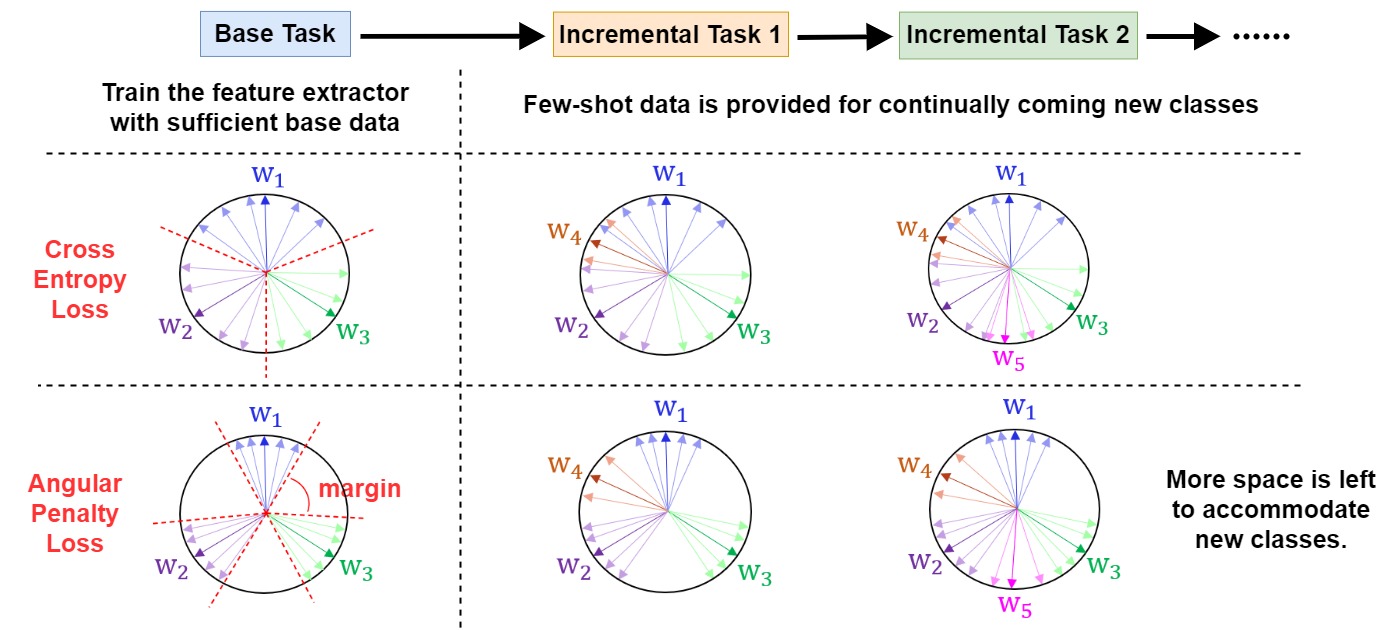}
	\caption{An illustration of feature distributions of a cross-entropy loss trained model and an angular penalty loss trained model.
		The light color arrows represent examples of different class features on the latent feature space.
		The dark color arrows represent the average feature prototype of corresponding classes.
		Angular penalty loss provides more compact intra-class clustering and wider inter-class separation than cross-entropy loss.
		Compact clustering leaves more room on the latent feature space to accommodate the new classes.
	}
	\label{fig: angular_penalty}		
\end{figure}

Under the FSCIL setting, we want to obtain a feature extractor which can rapidly adapt to continually coming new tasks, as well as be stable to overcome catastrophic forgetting for the previously learned tasks.
Thus, we want to use a loss function that:
1) minimizes the distance between intra-class feature vectors, and
2) maximizes the distance between inter-class feature vectors.
The compact intra-class clustering and wide inter-class separation will leave more room in the latent feature space for the incrementally arriving new classes and hence lead to better open-set classification.
Figure \ref{fig: angular_penalty} illustrates an example.
As many innovative angular penalty losses have been explored and proposed for face recognition studies \cite{wang2018cosface,deng2019arcface} and considering the similarity between FSCIL and face recognition tasks, we adapt the cosFace penalty strategy \cite{wang2018cosface} to FSCIL training. 

First, we use cosine similarity as the distance metric to measure data similarity and compute scores.
It has two effects:
1) it makes training focus on the angles between normalized features instead of absolute distance in the latent feature space, and
2) the normalized weight parameters of the fully connected layer can be regarded as the center of each category.
To calculate cosine similarity in the final fully connected layer, we fix the bias to 0 for simplicity.
Then the data prediction procedure can be written as:
\begin{small}
	\begin{equation}
	f = \mathcal F(x)
	\label{eq: feature_extractor}
	\end{equation}
\end{small}
\begin{small}
	\begin{equation}
	y_i = W_i^T  f = \|W_i\| \|f\| \cos(\theta_i) = \cos(\theta_i), \nonumber
	\end{equation}
	\begin{equation}
	\|W_i\| = \|f\| = 1
	\end{equation}
	\label{eq: cosine_similarity}
\end{small}

\noindent where $f$ is the feature obtained from the input image $x$ through the feature extractor $\mathcal F$.
The feature $f$ and the weight parameter $W_i$ are normalized by $\ell 2$ normalization, so the magnitude is 1.
The quantity $y_i$ is the calculated cosine similarity between the feature $f$ and the weight parameter $W_i$ for class $i$.
It measures the angular similarity of image $x$ towards class $i$ which indicates the likelihood that image $x$ belongs to class $i$.

Normally, the cosine similarity prediction is used with cross-entropy loss to separate features from different classes by maximizing the probability of the ground-truth class.
The loss function is:
\begin{small}
	\begin{equation}
	\begin{aligned}
	L & = -\frac{1}{N}\sum_{j=1}^{N}\log(p_j) = -\frac{1}{N}\sum_{j=1}^{N}\log(\frac{e^{y_{j}}}{\sum^{C}_{i=1}e^{y_i}}),\\
	& = -\frac{1}{N}\sum_{j=1}^{N}\log(\frac{e^{\|W_{j}\| \|f\| \cos(\theta_{j})}}{\sum_{i=1}^{C}e^{\|W_i\| \|f\| \cos(\theta_i)}}),\\
	& = -\frac{1}{N}\sum_{j=1}^{N}\log(\frac{e^{\cos(\theta_{j})}}{\sum_{i=1}^{C}e^{\cos(\theta_i)}}) \\
	\end{aligned}
	\label{eq: cross_entropy}
	\end{equation}
\end{small}

\noindent where $N$ is the number of training images and $C$ is the number of classes.
The quantity $p_j$ describes the softmax probability for image $j$.
The quantity $y_j$ describes the cosine similarity towards its ground truth class for image $j$.

To make features better clustered, inspired by cosFace \cite{wang2018cosface}, a cosine margin $m$ is introduced to the classification boundary.
With the help of the extra margin, the intra-class features become more compactly clustered and the inter-class features become more widely separated.
Following cosFace, we also re-scale the normalized feature by a preset scale factor $s$.
The loss function is:
\begin{small}
	\begin{equation}
	\begin{split}
	L_{AP} = -\frac{1}{N}\sum_{j=1}^{N}\log(\frac{e^{s(\cos(\theta_{j}) - m)}}{e^{s(\cos(\theta_{j}) - m)} + \sum_{i\neq j}e^{s\cos(\theta_i)}})
	\label{eq: angular_penalty}
	\end{split}
	\end{equation}
\end{small}
\noindent The scale factor $s$ is set to 30 and the cosine margin $m$ is set to 0.4 for all experiments.

\subsection{Augmented Training}
Diverse and transferable representation is the key for open-set problems.
Exposure to a large number of classes is one way to obtain such kind of feature extractors.
To this end, a simple and effective method is to introduce auxiliary classes. 
\begin{wrapfigure}{r}{0.5\textwidth}
	\centering
	\includegraphics[width=0.47\textwidth]{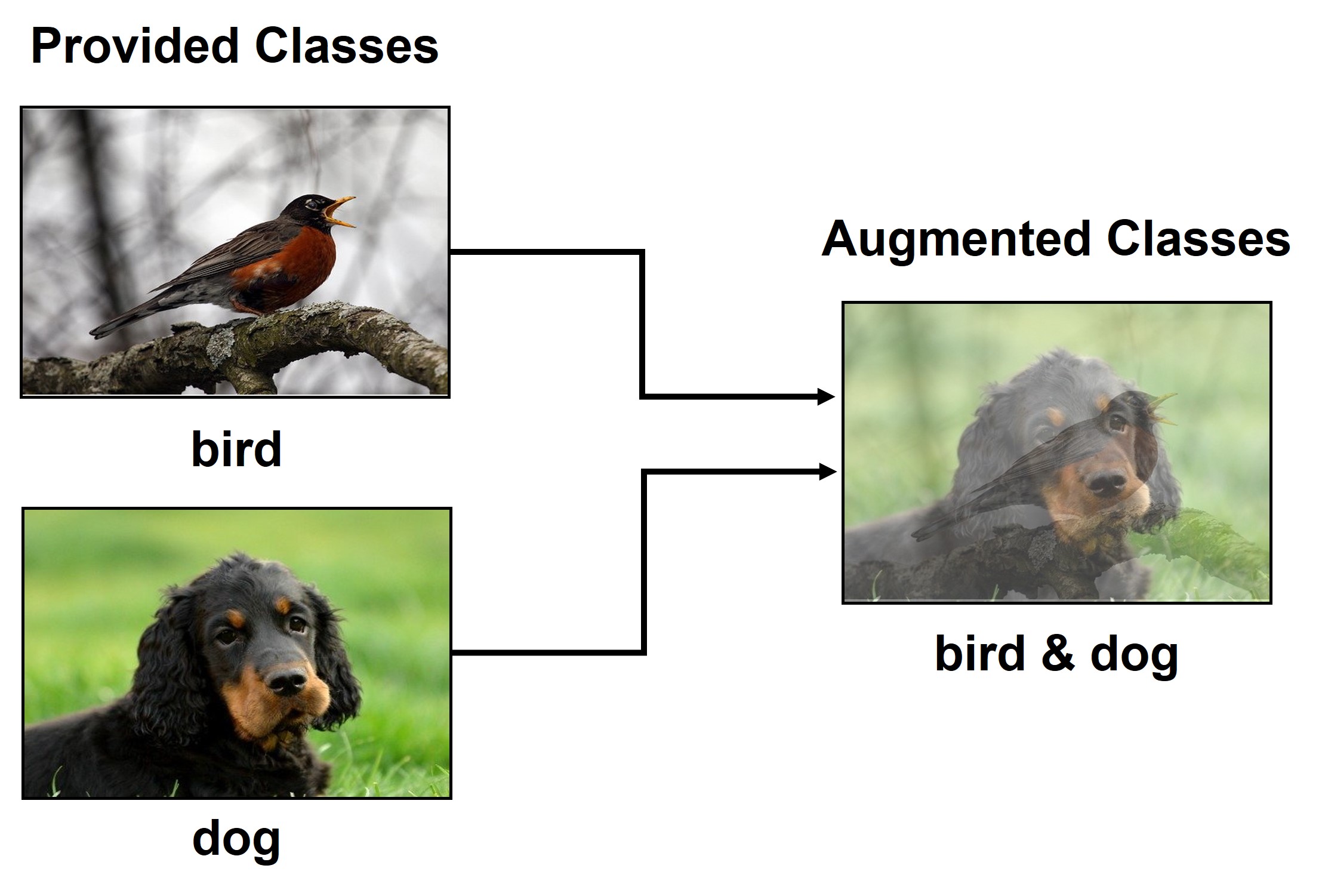}	
	\caption{An example of class augmentation. 
		Auxiliary new class data is generated by interpolating two different class samples from base session data.}
	\label{fig: class_fusion}		
\end{wrapfigure}
Inspired by Mixup \cite{zhang2017mixup} and IL2A \cite{zhu2021class}, we randomly combine pairs of different class examples from the base session data to synthesize auxiliary new class data.
The new class data generating function is:
\begin{small}
	\begin{equation}
	\begin{split}
	x_{k} = \lambda x_i + (1 - \lambda) x_j
	\label{eq: class_augmentation}
	\end{split}
	\end{equation}
\end{small}
\noindent where $x_i$ and $x_j$ are two training samples from two different classes $i$ and $j$ randomly picked from the $C$ base session classes.
$\lambda$ is the interpolation coefficient.
$x_{k}$ is the generated new class data.
Figure \ref{fig: class_fusion} shows an example.
In our experiments, following IL2A \cite{zhu2021class}, we restrict $\lambda$ to be a randomly chosen value between $[0.4, 0.6]$ to reduce the overlap between the augmented and original classes. 
For a $C$-class classification task, by pair combination, we will generate $(C \times (C - 1) / 2)$ new classes, so the original $C$-class classification task now becomes a $(C + C \times (C - 1) / 2)$-class classification task.

Exposure to various image conditions during training is also a good method to obtain a general feature extractor.
Inspired by self-supervised learning \cite{chen2020simple,chen2021exploring}, we use two augmentations of each image to enhance training data diversity.
Figure \ref{fig: framework} shows the augmentation procedure.
During training, for each input image, we randomly generate two augmentations from a set of preset transformation strategies.
For the utilized transformation methods, we randomly apply resized crop, horizontal flip, color jitter, and grayscale.
Then both transformed data are sent to the backbone network.
The losses from two sets of augmentation are averaged and back-propagated to update model parameters.
In addition, to avoid the feature extractor over-specialize to base session data, following SimCLR \cite{chen2020simple}, we utilize extra projection layers before the final fully connected layer.
By leveraging the nonlinear projection head, more information can be formed and maintained in the feature extractor. 

\subsection{Balanced Testing}
After base session training, the projection head and the augmented classification head are discarded.
Only the feature extractor is left and it is frozen to avoid both overfitting and catastrophic forgetting.
During testing, nearest class mean and cosine similarity are utilized to do the classification.
As there is only limited data provided for each incremental session, to alleviate the possible prediction bias due to data imbalance, we use the same amount of few-shot data as the following incremental steps to generate the base class prototypes.
To select suitable examples, we first use all base session data to calculate the class-wise mean for each base class.
Then the required few-shot amount of data which has the smallest cosine distance with the calculated mean is used to generate the final prototype for each base session class.

\subsection{Harmonic Accuracy}
For the evaluation metric, current SOTA methods generally report the class-wise average accuracy.
However, we argue that the class-wise average accuracy is not enough to evaluate the performance of an FSCIL method, since the number of classes from the base session is often a large fraction of the total number of classes.
Following the experimental settings on benchmark papers, for CIFAR100 \cite{krizhevsky2009learning} and miniImageNet \cite{russakovsky2015imagenet}, 60 out of 100 (60\%) categories are used as base classes.
For CUB200 \cite{wah2011caltech}, 100 out of 200 (50\%) categories are used as base classes.
A model with good performance on the base session and poor performance on the following incremental sessions can still have a good average accuracy due to the high ratio of base classes to the overall classes. 
For example, with 60 base classes, on one step of incremental learning 5 classes, an algorithm that shows 100\% accuracy on base classes with 0\% on incremental classes would be rated 92.3\% using average accuracy, yet it would have demonstrated no learning on the new task.
To compensate for this deficiency of average accuracy, we adapt the harmonic accuracy metric that requires well-balanced performance across both base and incremental classes.  
The formula for harmonic accuracy ($A_{h}$) is:
\begin{small}
	\begin{equation}
	\begin{split}
	A_{h} = \frac{2 \times A_{b} \times A_{i}}{A_{b} + A_{i}}
	\label{eq: harmonic_accuracy}
	\end{split}
	\end{equation}
\end{small}

\noindent where $A_{b}$ is the average accuracy for base session classes and $A_{i}$ is the average accuracy for the following incremental session classes.
In the simple example above, the harmonic accuracy would be 0\% which is much more appropriate as the network has indeed learned nothing at all. 
An ideal balanced FSCIL classifier will have equally high performance on both average accuracy and harmonic accuracy.
If a model has good average accuracy but poor harmonic accuracy, this means that its good performance is mainly due to performance on the base session classes and the model has poor incremental learning capability overall.

\section{Experiments}
\label{sec: Experiments}

\subsection{Dataset and Evaluation Metric}
We use three benchmark datasets CIFAR100 \cite{krizhevsky2009learning}, miniImageNet \cite{russakovsky2015imagenet} and Caltech-UCSD Birds-200-2011 (CUB200) \cite{wah2011caltech} for our experiments.
CIFAR100 contains 100 classes with 600 images per class, 500 for training and 100 for testing.
Each image has a size of 32 $\times$ 32 pixels.
MiniImageNet also contains 100 classes with 600 images per class, 500 for training and 100 for testing.
Each image has a size of 84 $\times$ 84 pixels.
CUB200 is a fine-grained image classification dataset.
It contains 200 classes of different species of birds with 5994 training images and 5794 testing images.
Each image has a size of 224 $\times$ 224 pixels.

As mentioned in section \ref{sec: Problem Formulation}, to comprehensively evaluate an FSCIL method, we follow the benchmark 5-shot setting and also perform an additional 1-shot setting.
For experiments on CIFAR100 and miniImageNet, the 8-step 5-way 5-short and 8-step 5-way 1-short incremental settings are used.
In this protocol, 60 classes are used as base classes with all training data provided; then
40 classes are used as incremental classes with 5-shot or 1-shot training data provided in a 5-way manner in 8 steps.
For experiments on CUB200, 10-step 10-way 5-shot and 10-step 10-way 1-shot settings are used.
100 classes are used as base classes and the remaining 100 classes are used as incremental classes with 5-shot or 1-shot training data provided in a 10-way manner in 10 steps.
To make the evaluation comprehensive and fair, we report both average accuracy and harmonic accuracy.

\subsection{Implementation Details}
For our experiments, we use ResNet18 \cite{he2016deep} as the backbone network.
We implement the projection head as a two-layer MLP with a hidden feature size of 2048 and ReLU as the activation function.
Our method is built with PyTorch library and SGD with momentum is used for optimization.
The initial learning rate is set to 0.01 for CIFAR100 and miniImageNet dataset training, and 0.001 for CUB200 dataset training.
Following the settings on \cite{tao2020few,zhang2021few}, models for CIFAR100 and miniImageNet are trained from scratch, and models for CUB200 are initialized by an ImageNet pretrained model.
When class augmentation is not applied, a batch size of 512 is used for training. 
When class augmentation is used, we use a batch size of 128 for CIFAR100 and a batch size of 64 for miniImageNet.
The experimental results for CEC \cite{zhang2021few} are reproduced by their publicly available source code. 

\begin{figure}[tbp]
	\centering
	\begin{subfigure}[t]{0.32\linewidth}
		\includegraphics[width=\textwidth,keepaspectratio]{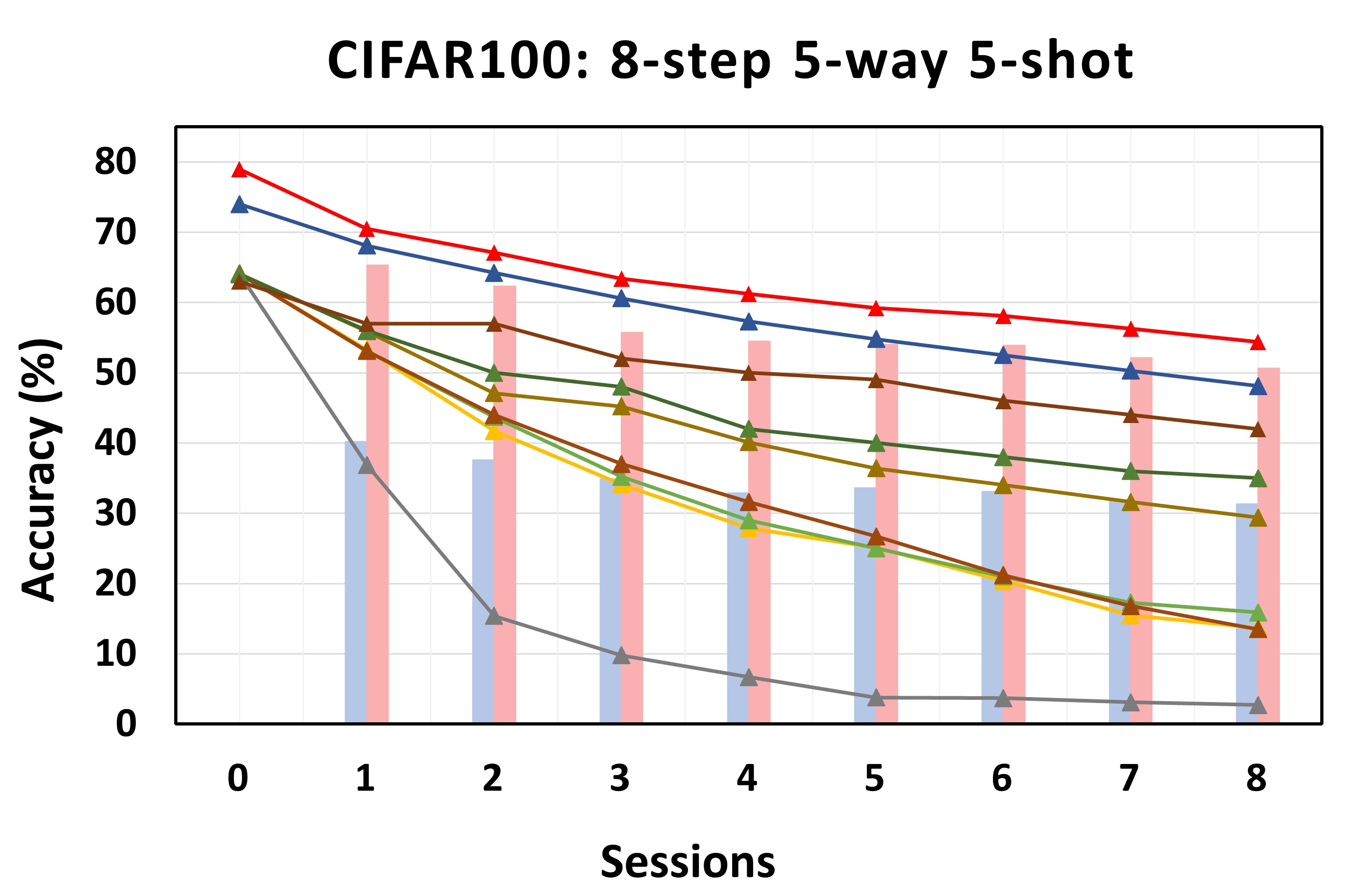}
		\label{fig:exp_cifar100_5shot}		
	\end{subfigure}
	\begin{subfigure}[t]{0.32\linewidth}
		\includegraphics[width=\textwidth,keepaspectratio]{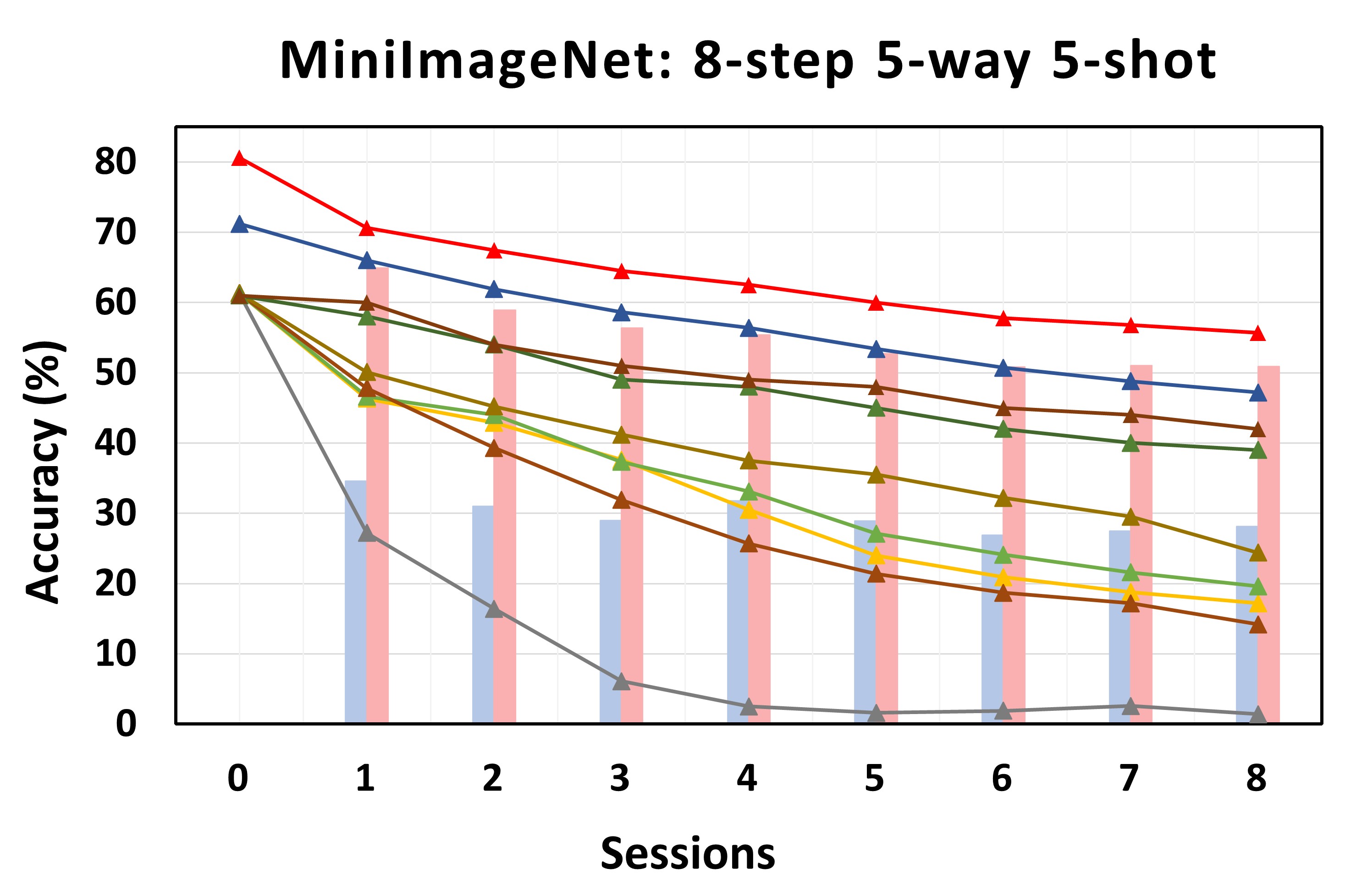}
		\label{fig:exp_miniimagenet_5shot}		
	\end{subfigure}
	\begin{subfigure}[t]{0.32\linewidth}
		\includegraphics[width=\textwidth,keepaspectratio]{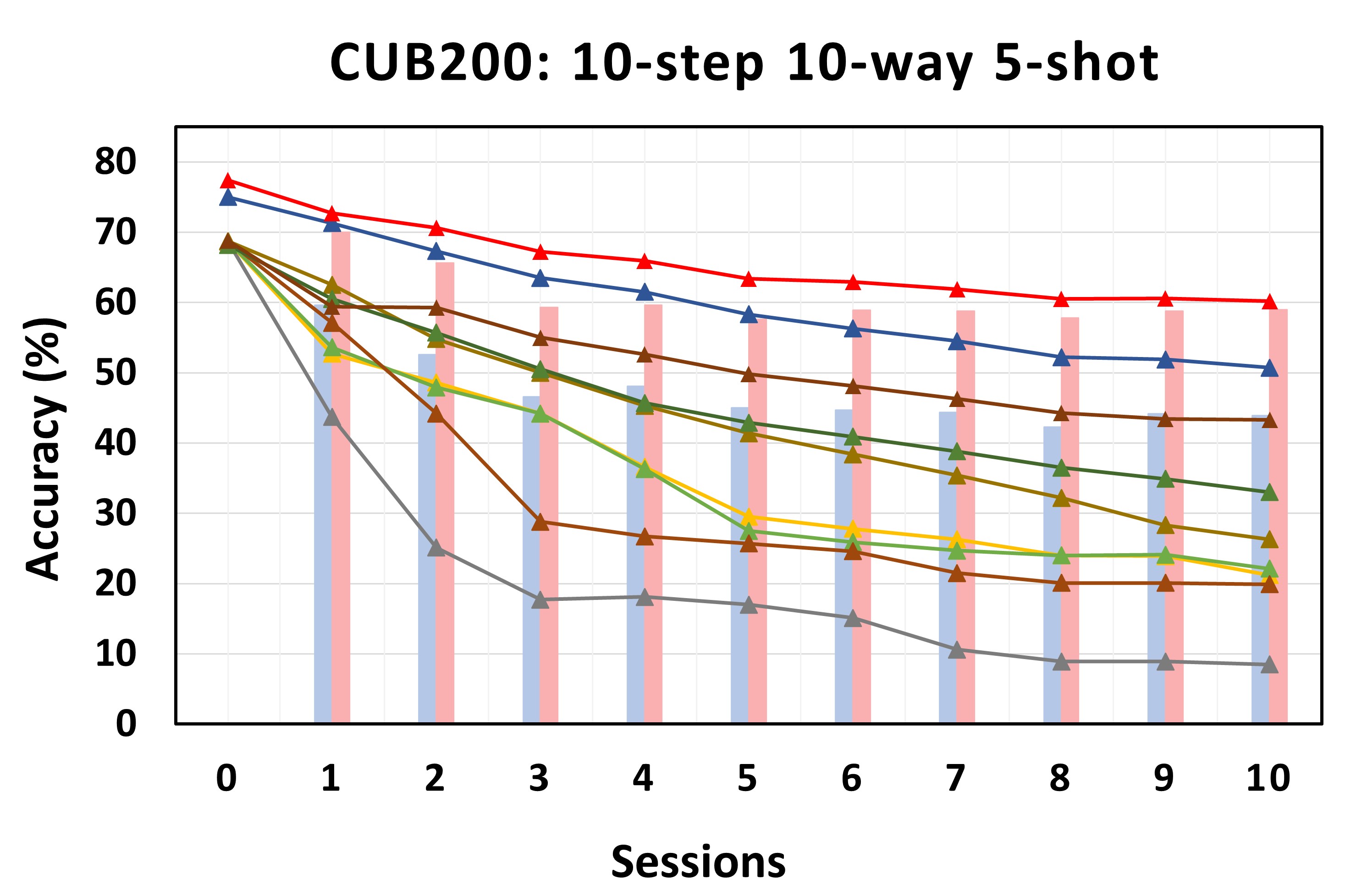}
		\label{fig:exp_cub200_5shot}		
	\end{subfigure}
	\begin{subfigure}[t]{0.32\linewidth}
		\includegraphics[width=\textwidth,keepaspectratio]{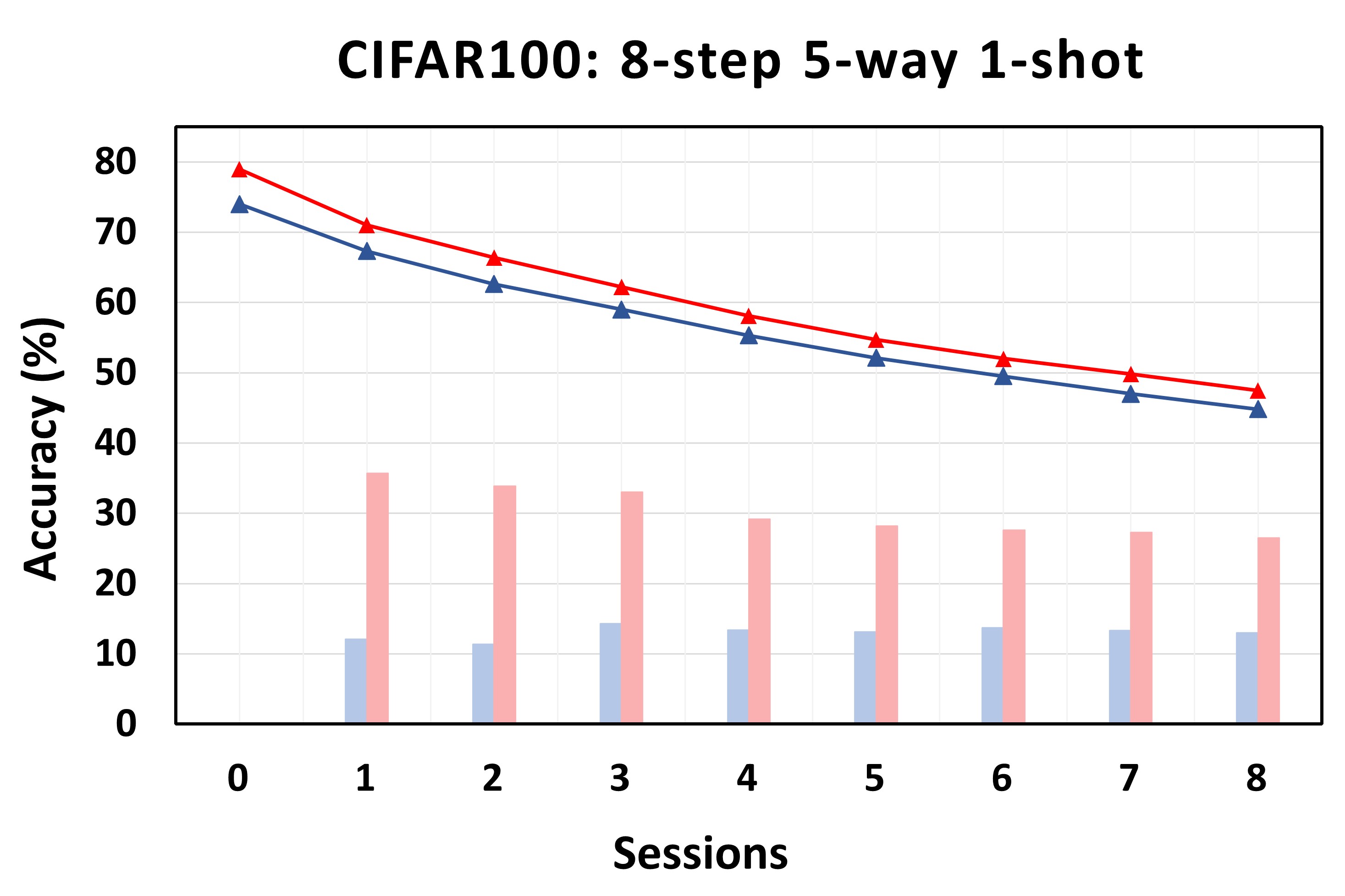}
		\label{fig:exp_cifar100_1shot}		
	\end{subfigure}
	\begin{subfigure}[t]{0.32\linewidth}
		\includegraphics[width=\textwidth,keepaspectratio]{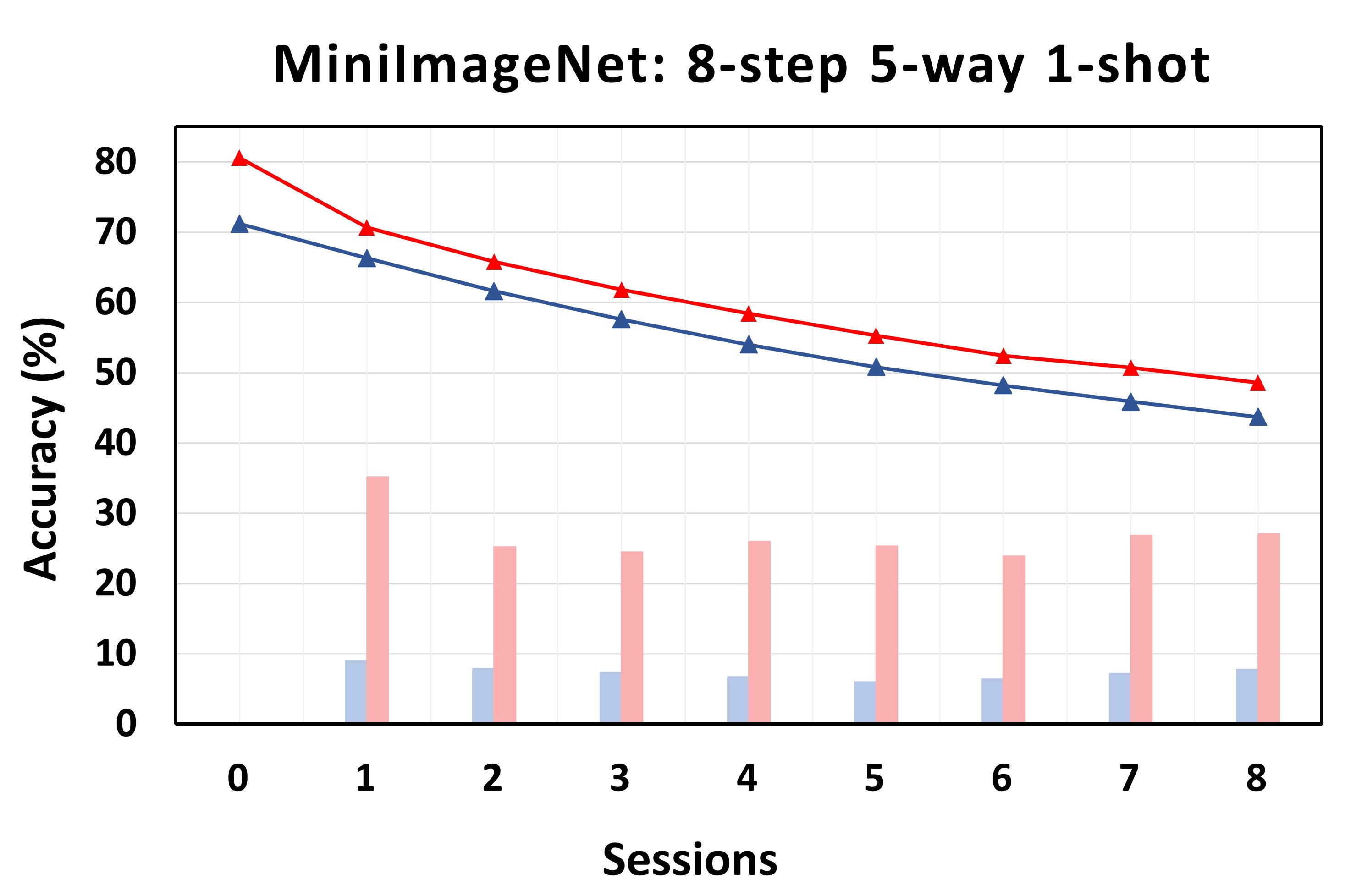}
		\label{fig:exp_miniimagenet_1shot}			
	\end{subfigure}
	\begin{subfigure}[t]{0.32\linewidth}
		\includegraphics[width=\textwidth,keepaspectratio]{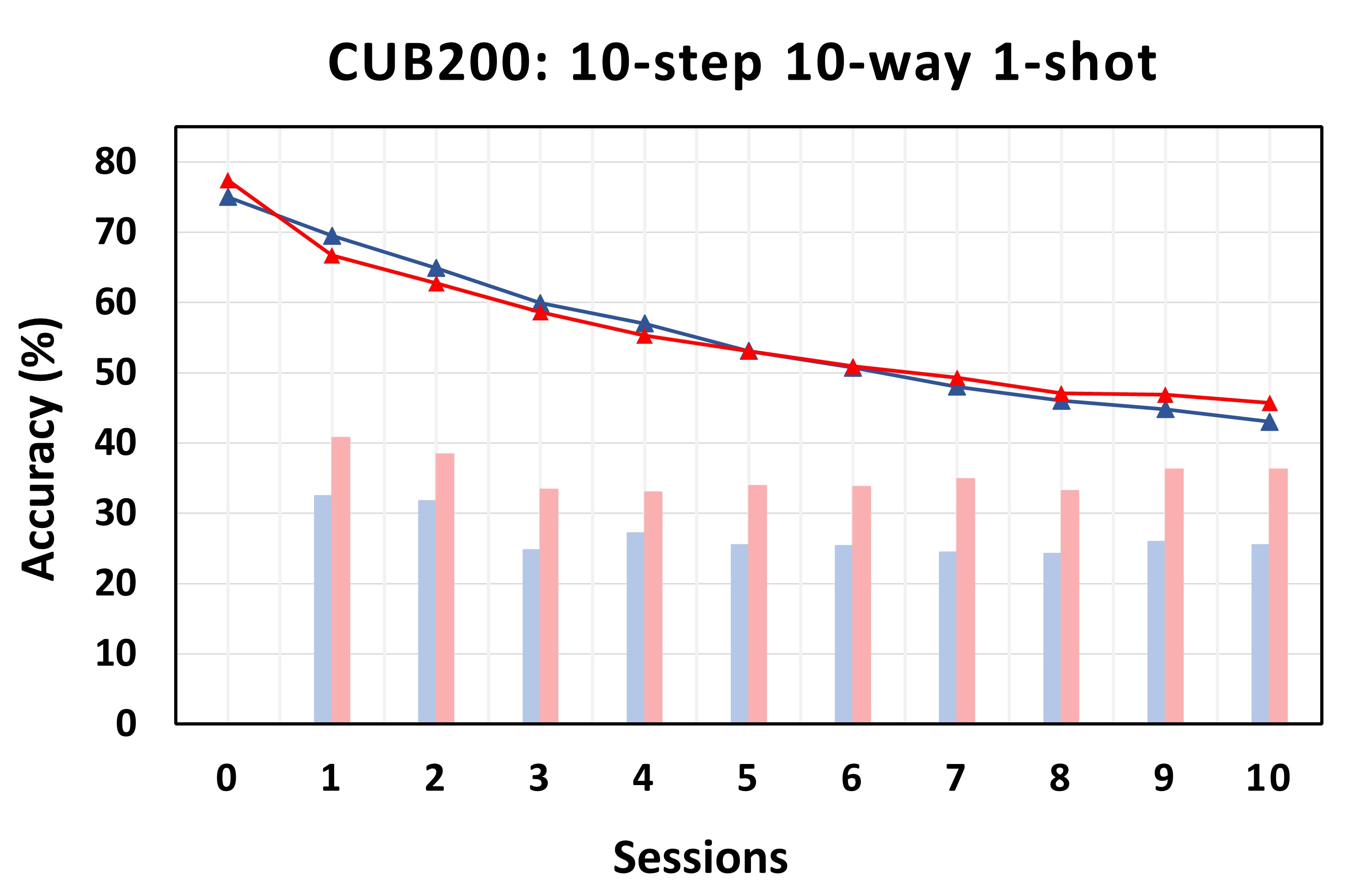}
		\label{fig:exp_cub200_1shot}			
	\end{subfigure}
	\begin{subfigure}[t]{0.9\linewidth}
		\includegraphics[width=\textwidth,keepaspectratio]{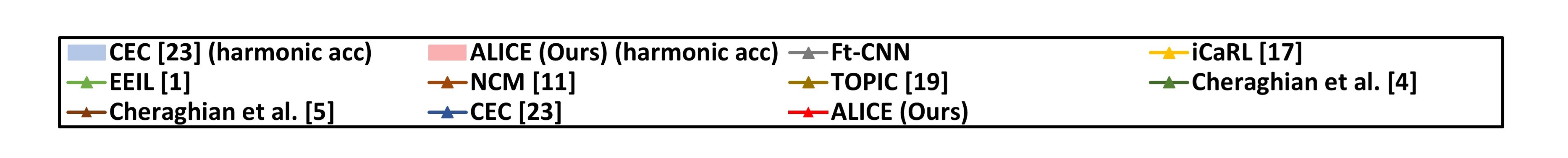}
		\label{fig:exp_caption_v1}	
	\end{subfigure}
	\caption{Comparison with SOTA methods under both 5-shot and 1-shot incremental settings on CIFAR100, miniImageNet, and CUB200 dataset.
		The line chart represents average accuracy and the histogram represents harmonic accuracy. 
		Our method outperforms SOTA works with significant performance advantages.}
	\label{fig:exp_cifar100_miniimagenet}
\end{figure}

\subsection{Comparison with the State-of-the-art Methods}
We compare our method with the SOTA methods \cite{rebuffi2017icarl,castro2018end,hou2019learning,tao2020few,cheraghian2021semantic,cheraghian2021synthesized,zhang2021few} on three datasets.
According to Figure \ref{fig:exp_cifar100_miniimagenet}, for experiments on CIFAR100 and miniImageNet dataset, under both 8-step 5-way 5-shot and 1-shot settings, our method achieves the highest class-wise accuracy over all the sessions.
Also, on both datasets, our ALICE method shows much higher harmonic accuracy on all sessions compared to the SOTA CEC \cite{zhang2021few} method.
The high harmonic accuracy proves that our method can largely alleviate the prediction bias problem.
To be more specific, in CIFAR100, for the 5-shot (1-shot) setting, in the last session, we get 54.1\% (47.5\%) average accuracy and 50.6\% (26.5\%) harmonic accuracy which is 6.0\% (2.7\%) and 19.3\% (13.5\%) higher than the CEC method, respectively. 
In miniImageNet, for the 5-shot (1-shot) setting, in the last session, we get 55.7\% (48.6\%) average accuracy and 50.9\% (27.1\%) harmonic accuracy which is 8.5\% (4.9\%) and 22.8\% (19.3\%) higher than the CEC method, respectively.

For experiments on the CUB200 dataset, we find that applying class augmentation will deteriorate the model performance.
As CUB200 is a fine-grained dataset, the feature extractor needs to focus on learning tiny differences between categories. 
However, class augmentation targets obscuring the class difference and forcing the feature extractor to focus on general features.
It is a good augmentation strategy to extract transferable features for general FSCIL tasks but will adversely obscure the class boundaries for fine-grained FSCIL tasks.
Thus, for experiments on CUB200, we do not use class augmentation and train the feature extractor only by angular penalty and data augmentation.
According to Figure \ref{fig:exp_cifar100_miniimagenet}, our method outperforms all SOTA methods by a large margin on the 5-shot setting.
This proves that for fine-grained classification where the re-usability of features is high, angular penalty and data augmentation is enough to obtain a robust open-set feature extractor. 
Under the 1-shot setting, we get similar average accuracy as CEC, since both of us freeze the backbone network after base session training to avoid catastrophic forgetting. 
When considering incremental class performance, our method can better adapt to new classes and obtain much higher harmonic accuracy than the CEC method.

\begin{figure}[tbp]
	\centering
	\begin{subfigure}[t]{0.22\linewidth}
		\includegraphics[width=\textwidth,keepaspectratio, trim={0cm 0cm 3.6cm 0cm}, clip]{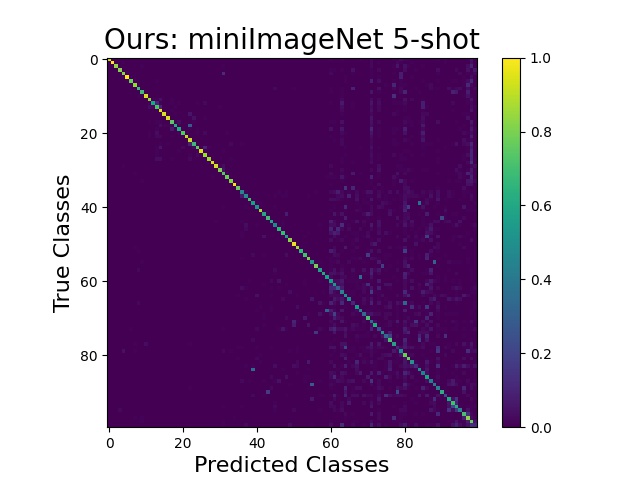}
		\label{fig: cm_miniimagenet_5shot_alice}		
	\end{subfigure}
	\begin{subfigure}[t]{0.22\linewidth}
		\includegraphics[width=\textwidth,keepaspectratio, trim={0cm 0cm 3.6cm 0cm}, clip]{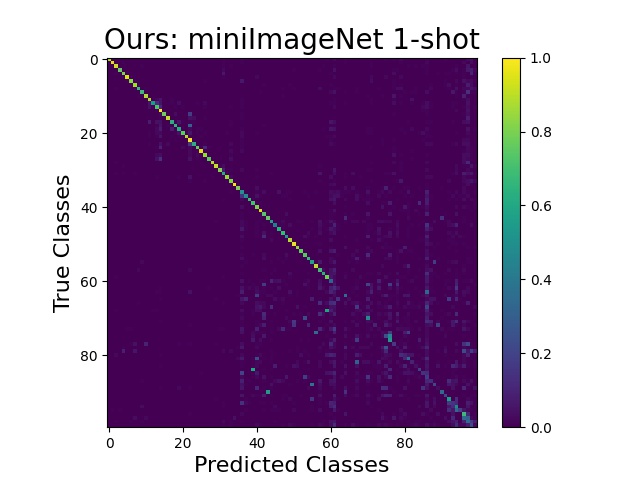}
		\label{fig: cm_miniimagenet_1shot_alice}		
	\end{subfigure}
	\begin{subfigure}[t]{0.22\linewidth}
		\includegraphics[width=\textwidth,keepaspectratio, trim={0cm 0cm 3.6cm 0cm}, clip]{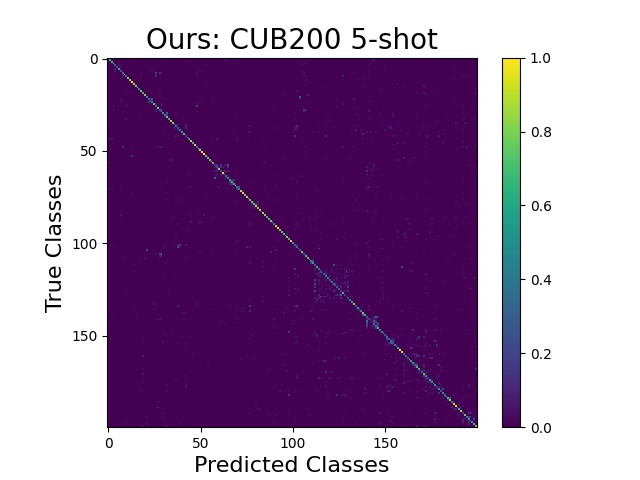}
		\label{fig: cm_miniimagenet_5shot_cec}
	\end{subfigure}
	\begin{subfigure}[t]{0.246\linewidth}
		\includegraphics[width=\textwidth,keepaspectratio, trim={0cm 0cm 2cm 0cm}, clip]{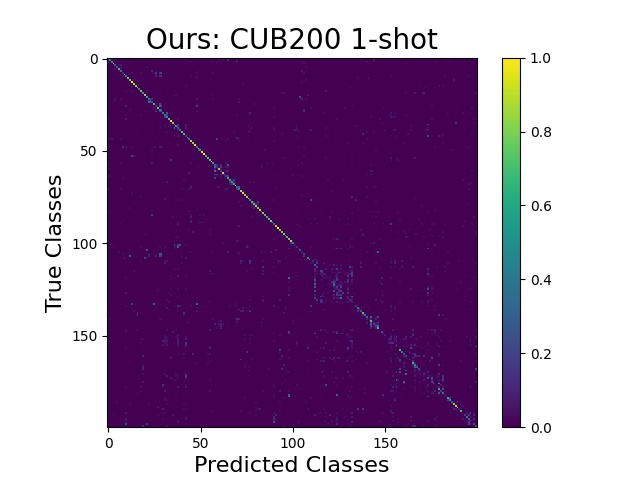}
		\label{fig: cm_miniimagenet_1shot_cec}	
	\end{subfigure}
	\begin{subfigure}[t]{0.22\linewidth}
		\includegraphics[width=\textwidth,keepaspectratio, trim={0cm 0cm 3.6cm 0cm}, clip]{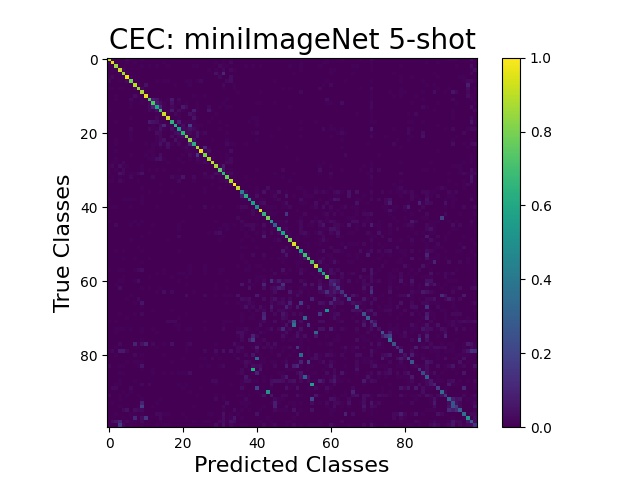}
		\label{fig: cm_cub_5shot_alice}	
	\end{subfigure}
	\begin{subfigure}[t]{0.22\linewidth}
		\includegraphics[width=\textwidth,keepaspectratio, trim={0cm 0cm 3.6cm 0cm}, clip]{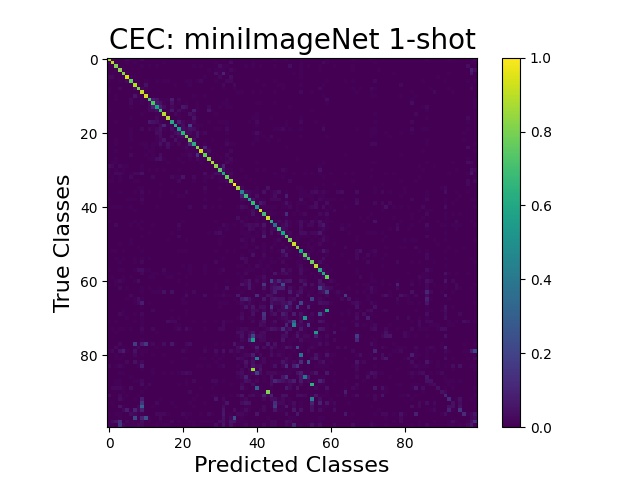}
		\label{fig: cm_cub_1shot_alice}	
	\end{subfigure}
	\begin{subfigure}[t]{0.22\linewidth}
		\includegraphics[width=\textwidth,keepaspectratio, trim={0cm 0cm 3.6cm 0cm}, clip]{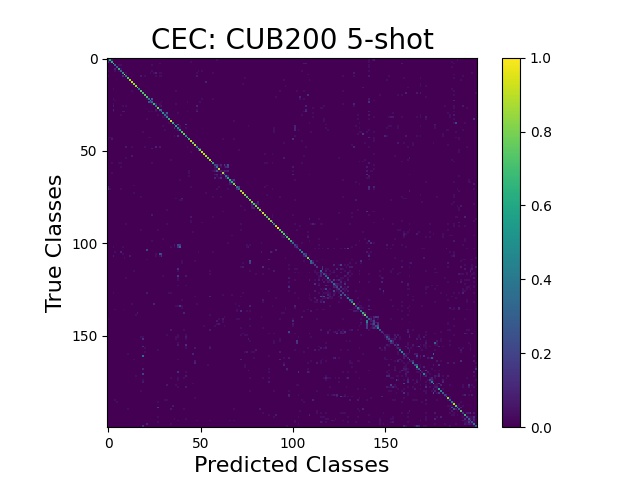}
		\label{fig: cm_cub_5shot_cec}	
	\end{subfigure}
	\begin{subfigure}[t]{0.246\linewidth}
		\includegraphics[width=\textwidth,keepaspectratio, trim={0cm 0cm 2cm 0cm}, clip]{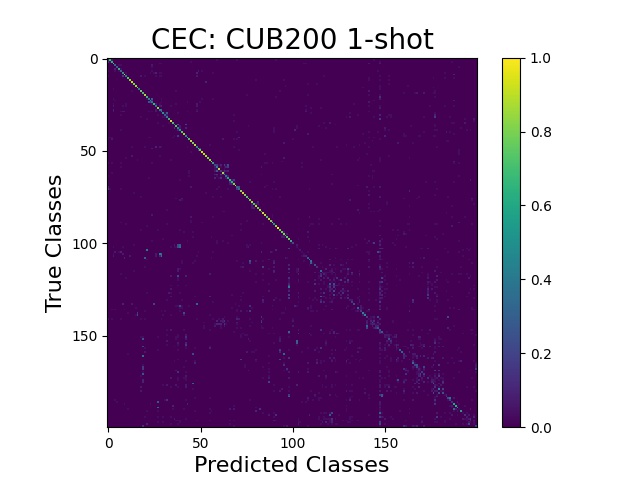}
		\label{fig: cm_cub_1shot_cec}	
	\end{subfigure}
	\caption{Comparison of the confusion matrices produced by CEC and our method on the last incremental session for 5-shot and 1-shot incremental experiments on miniImageNet and CUB200 dataset.}	
	\label{fig: cm}
\end{figure}

Besides, we also compare the confusion matrices produced by CEC and our method after the last incremental session.
The results are shown in Figure \ref{fig: cm}.
Compared with CEC, our method produces a more balanced base and incremental class performance, especially under 5-shot settings.
When under 1-shot settings, although our method can outperform the CEC method, the prediction bias towards base classes still exists.
This is because the 1-shot setting is the most extreme FSCIL setting due to maximal data scarcity and data imbalance. 
We will focus on solving this problem in future work.

\subsection{Ablation Study}
\label{sec: Ablation Study}
To validate the effectiveness of each part of our method, we perform an ablation study on the CIFAR100 dataset under the 8-step 5-way 5-shot setting.
Table \ref{tab: ablation} shows the experimental results for the ablation study.
When balanced data are used for prototype generation, compared with the cross-entropy loss trained model, in the last incremental step, the angular penalty loss trained model provides 3.3\% average accuracy improvement.
But the harmonic accuracy is 6.2\% lower.
This means that solely using angular penalty loss will make the feature extractor over-specialize to the base session data and lose its generalization performance.
The high average accuracy produced by the angular penalty loss trained model with all data used for prototype generation also shows the over-specialization.
Their good average performance is mainly due to the base classes since at the first several sessions, the ratio of base classes among all classes is high.
To compensate for the loss of generalization, projection layers are utilized to help the feature extractor maintain more information.
With the help of the projection layers, in the last incremental session, the average accuracy remains unchanged but the harmonic accuracy is increased from 25.2\% to 43.8\% which is 12.4\% higher than the cross-entropy loss trained model. 
Then, when two transformations of each input image are utilized for loss calculation, the average accuracy (harmonic accuracy) is 1.8\% (3.6\%) increased in the last step. 
After that, when class augmentation is applied, the average accuracy (harmonic accuracy) is increased by 2.8\% (3.2\%) in the last step. 
In addition, when balanced data is used for prototype generation, the harmonic accuracy for both cross-entropy and angular penalty loss trained model is increased.
This shows that simply utilizing the same amount of data from the base and incremental sessions to generate class prototypes can effectively alleviate the prediction bias due to data imbalance. 

\begin{table*}[tpb]
	\centering
	\scriptsize{
		\caption{Ablation study on CIFAR100 under the 8-step 5-way 5-shot setting.
		}
		\begin{tabular}{c|c c c c|c@{\hskip 0.28cm}c@{\hskip 0.28cm}c@{\hskip 0.28cm}c@{\hskip 0.28cm}c@{\hskip 0.28cm}c@{\hskip 0.28cm}c@{\hskip 0.28cm}c@{\hskip 0.28cm}c@{\hskip 0.28cm}}
			loss type & \tabincell{c}{class \\ aug} & \tabincell{c}{data \\ aug} & \tabincell{c}{project \\ layer} & \tabincell{c}{balanced \\ data} & 0 & 1 & 2 & 3 & 4 & 5 & 6 & 7 & 8 \\
			\hline
			& & & & & \multicolumn{9}{c}{class-wise average accuracy} \\
			\hline
			\multirow{2}*{\tabincell{c}{cross \\ entropy}} & \XSolid & \XSolid & \XSolid & \XSolid & 74.2 & 67.4 & 63.4 & 59.4 & 55.9 & 53.2 & 51.2 & 49.0 & 46.9 \\
			& \XSolid & \XSolid & \XSolid & \Checkmark & 74.2 & 65.4 & 61.6 & 57.7 & 54.5 & 52.1 & 49.9 & 47.9 & 46.2 \\
			\hline									
			\multirow{5}*{\tabincell{c}{angular \\ penalty}} & \XSolid & \XSolid & \XSolid & \XSolid & 76.9 & \textbf{72.9} & \textbf{68.2} & \textbf{64.1} & 60.3 & 57.0 & 54.3 & 51.9 & 49.7 \\
			& \XSolid & \XSolid & \XSolid & \Checkmark & 76.9 & 72.8 & 68.0 & 63.8 & 60.2 & 56.8 & 54.1 & 51.8 & 49.5 \\																	
			& \XSolid & \XSolid & \Checkmark & \Checkmark & 74.2 & 67.1 & 63.7 & 59.9 & 56.8 & 54.1 & 52.8 & 51.1 & 49.5 \\
			& \XSolid & \Checkmark & \Checkmark & \Checkmark & 75.6 & 68.2 & 64.2 & 60.3 & 57.9 & 55.6 & 54.7 & 53.1 & 51.3 \\				
			& \Checkmark & \Checkmark & \Checkmark & \Checkmark & \textbf{79.0} & 70.5 & 67.1 & 63.4 & \textbf{61.2} & \textbf{59.2} & \textbf{58.1} & \textbf{56.3} & \textbf{54.1} \\								
			\hline
			& & & & & \multicolumn{9}{c}{harmonic accuracy} \\
			\hline
			\multirow{2}*{\tabincell{c}{cross \\ entropy}} & \XSolid & \XSolid & \XSolid & \XSolid & - & 36.5 & 32.1 & 29.4 & 27.1 & 27.2 & 28.3 & 27.2 & 27.4 \\
			& \XSolid & \XSolid & \XSolid & \Checkmark & - & 45.5 & 37.8 & 34.7 & 32.4 & 32.8 & 32.1 & 30.8 & 31.4 \\
			\hline									
			\multirow{5}*{\tabincell{c}{angular \\ penalty}} & \XSolid & \XSolid & \XSolid & \XSolid & - & 34.0 & 28.2 & 26.4 & 23.6 & 23.3 & 22.6 & 22.1 & 22.3 \\
			& \XSolid & \XSolid & \XSolid & \Checkmark & - & 40.4 & 32.8 & 29.7 & 27.5 & 26.2 & 25.4 & 24.6 & 25.2 \\							
			& \XSolid & \XSolid & \Checkmark & \Checkmark & - & 58.9 & 57.2 & 50.5 & 47.9 & 46.4 & 46.4 & 45.0 & 43.8 \\
			& \XSolid & \Checkmark & \Checkmark & \Checkmark & - & 65.0 & 60.0 & 52.2 & 50.9 & 49.6 & 50.1 & 48.6 & 47.4 \\					
			& \Checkmark & \Checkmark & \Checkmark & \Checkmark & - & \textbf{65.3} & \textbf{62.3} & \textbf{55.7} & \textbf{54.5} & \textbf{54.0} & \textbf{53.9} & \textbf{52.1} & \textbf{50.6} \\
		\end{tabular}	
		\label{tab: ablation}
	}
\end{table*}
\begin{figure}[tbp]
	\centering
	\begin{subfigure}[t]{0.3\linewidth}
		\centering
		\includegraphics[width=\textwidth,keepaspectratio, trim={1.3cm 1.3cm 1.3cm 1.3cm}, clip]{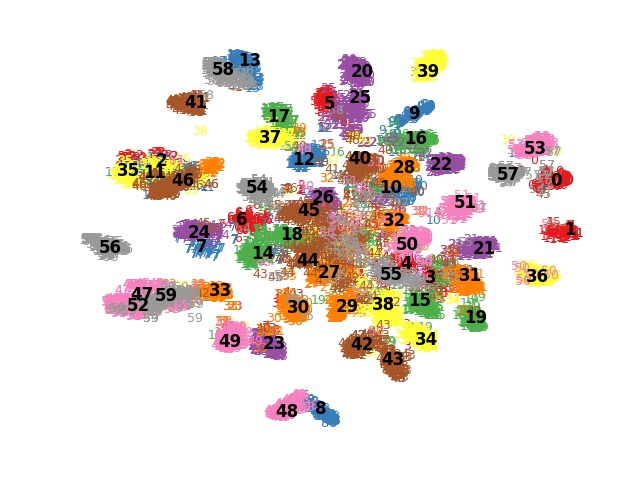}
		\caption{Feature distribution trained by cross-entropy loss.}
		\label{fig:CrossEntropy_trainset_feature_cls-0-60}		
	\end{subfigure}
	\quad
	\begin{subfigure}[t]{0.3\linewidth}
		\centering
		\includegraphics[width=\textwidth,keepaspectratio, trim={1.3cm 1.3cm 1.3cm 1.3cm}, clip]{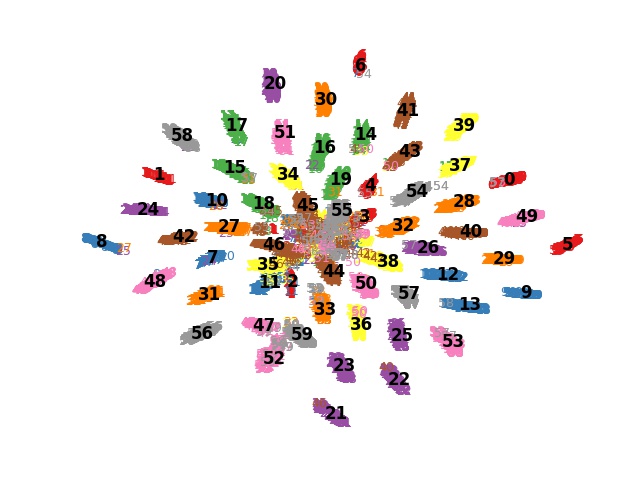}	
		\caption{Feature distribution trained by angular penalty loss.}
		\label{fig:AngularPenalty_trainset_feature_cls-0-60}		
	\end{subfigure}
	\quad
	\begin{subfigure}[t]{0.3\linewidth}
		\centering
		\includegraphics[width=\textwidth,keepaspectratio, trim={1.3cm 1.3cm 1.3cm 1.3cm}, clip]{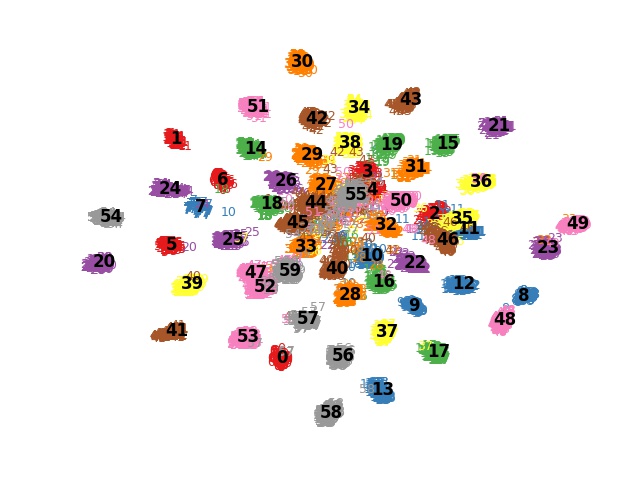}	
		\caption{Trained by angular penalty loss with projection layers. 
			Class and data augmentation are applied.}
		\label{fig:AngularPenalty_DA_trainset_feature_cls-0-60}			
	\end{subfigure}
	\caption{t-SNE visualization of the feature embeddings for the 60 base classes on CIFAR100.
		Each small colored number represents one feature instance for that class.
		The bold black number represents the average prototype for the class. 
	}
	\label{fig: ablation}
\end{figure}

Figure \ref{fig: ablation} shows the t-SNE visualization of the training data feature generated by different training strategies.
The model trained via angular penalty loss makes the training data cluster better in the latent feature space than the model trained via cross-entropy loss.
Then with the further help of projection layers, class and data augmentation, diverse and transferable features are obtained while different class features are still well separated.

\subsection{Hyper-parameter Analysis}
\label{sec: Hyper-parameter Analysis}
\begin{figure}[tbp]
	\centering
	\begin{subfigure}[t]{0.49\linewidth}
		\includegraphics[width=\textwidth,keepaspectratio]{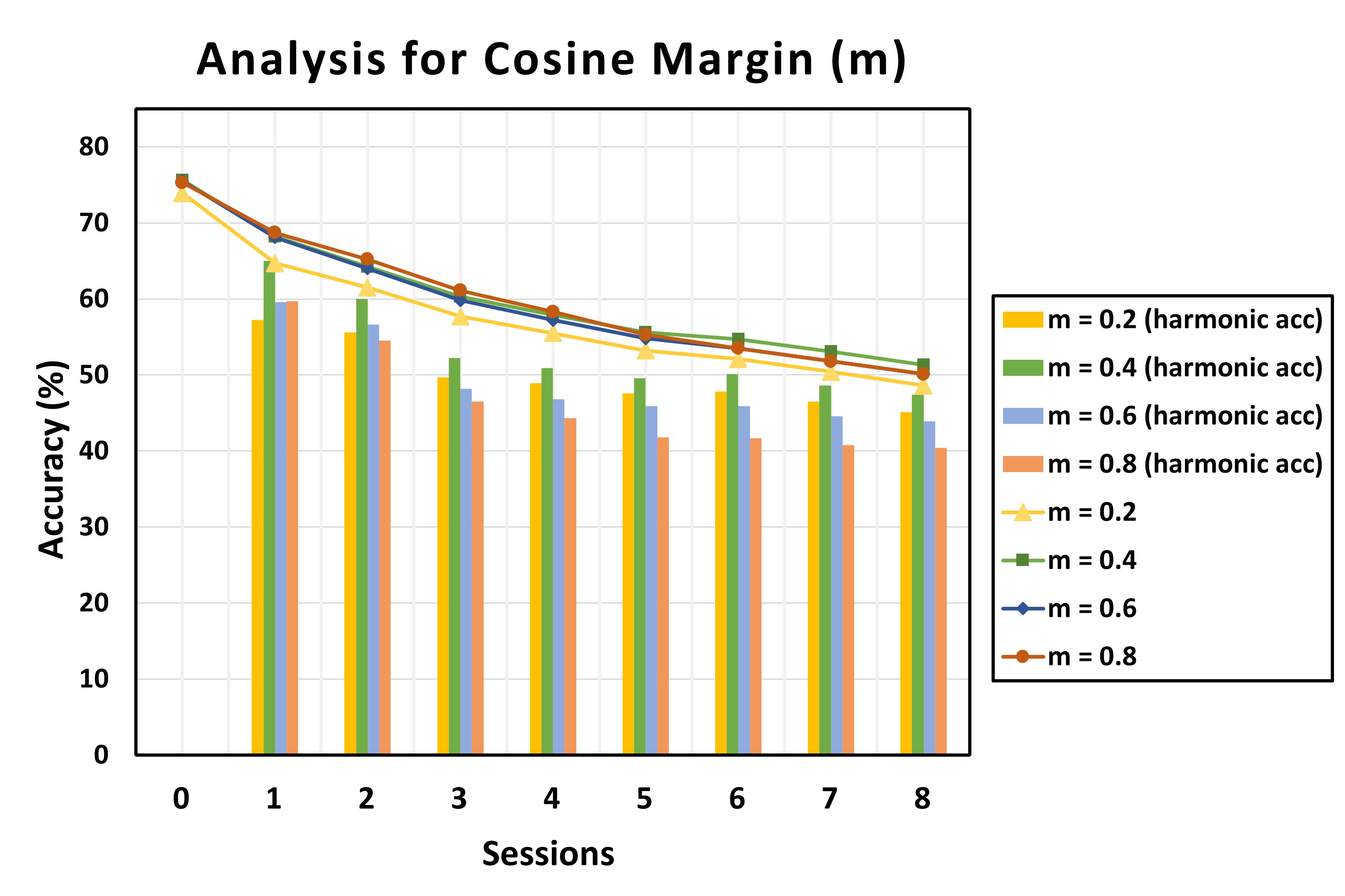}
		\label{fig:ablation_m}		
	\end{subfigure}
	\begin{subfigure}[t]{0.49\linewidth}
		\includegraphics[width=\textwidth,keepaspectratio]{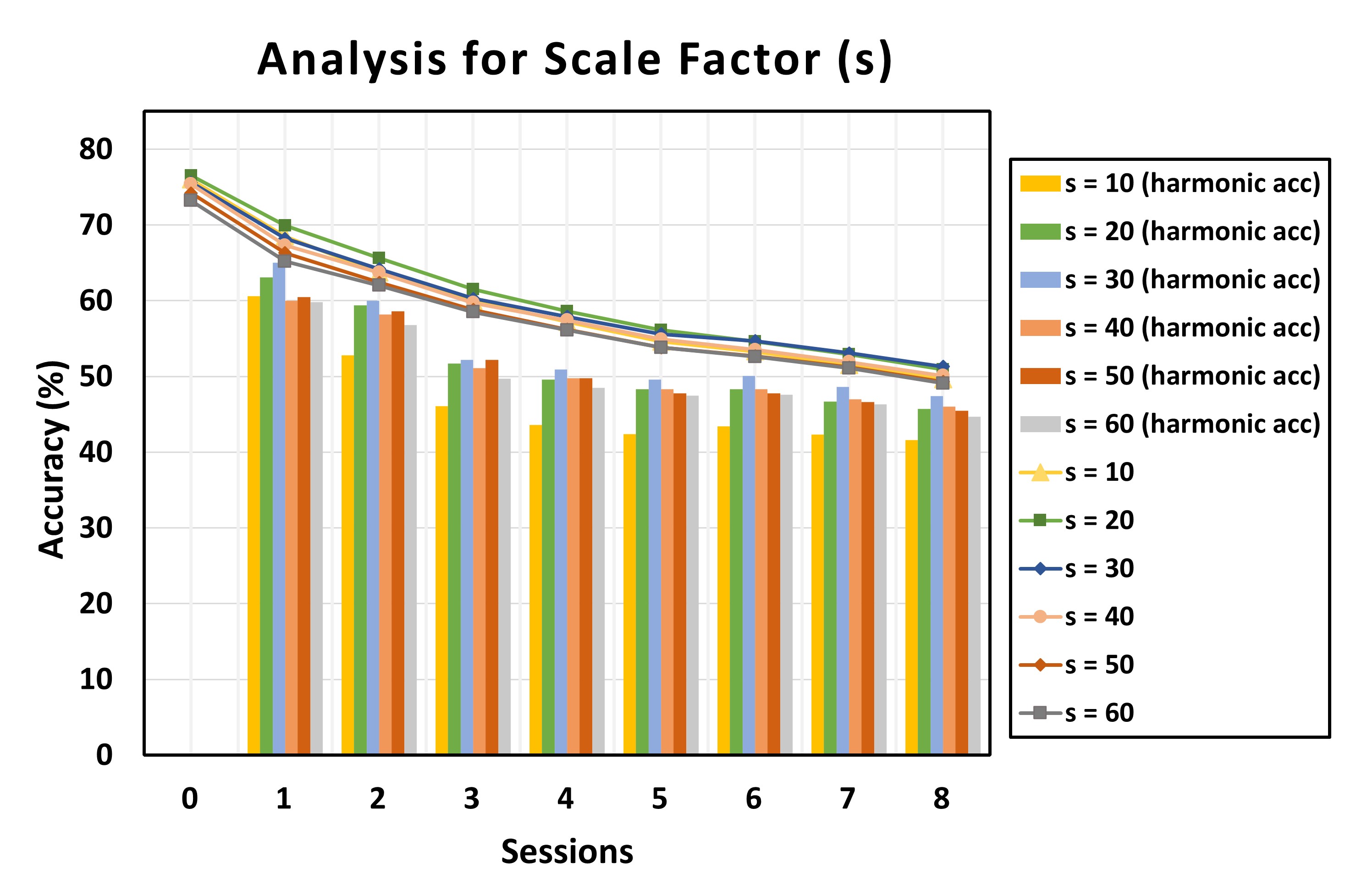}
		\label{fig:ablation_s}		
	\end{subfigure}
	\caption{Hyper-parameter studies for cosine margin ($m$) and scale factor ($s$) on CIFAR100 under the 8-step 5-way 5-shot setting.
		\label{fig:hyper_parameter}		
	}
\end{figure}

For the angular penalty loss calculation, there are two hyper-parameters --- the cosine margin ($m$) and the scale factor ($s$).
To find the most suitable hyper-parameter value, we perform the hyper-parameter grid analysis on the CIFAR100 dataset under the 8-step 5-way 5-shot protocol. 
All the experiments for hyper-parameter analysis are trained using angular penalty loss with data augmentation.
Figure \ref{fig:hyper_parameter} shows the experimental results.
First, we set the scale factor to 30 and vary the value for the cosine margin.
According to Figure \ref{fig:hyper_parameter}, we find that when the cosine margin is set to 0.4, in most sessions, the best average and harmonic accuracy are acquired.
Then, we set the cosine margin to 0.4 and vary the value for the scale factor.
We find that when the scale factor is set to 20 or 30, a good performance is usually acquired in most sessions.
Thus, for all our experiments, we set the cosine margin to 0.4 and the scale factor to 30.

\section{Conclusion}
\label{sec: Conclusion}
In this paper, we first reformulate the FSCIL task and propose a more practical and comprehensive setup. 
After that, inspired by techniques from modern face recognition and data augmentation, we proposed our ALICE method.
We link the relationship between FSCIL and open-set tasks and emphasize the importance of using base session training to obtain generalizable features for the FSCIL task.
We show that with only balanced nearest class mean and no further action in prototype evolution, our method outperforms the SOTA methods by substantial improvements in all benchmark datasets.

\noindent\textbf{Acknowledgments.}
We thank Dr. Yadan Luo and Kaiyu Guo for their help, discussion, and support.
This research was funded by the Australian Government through the Australian Research Council and Sullivan Nicolaides Pathology under Linkage Project LP160101797.

\clearpage
%
%
\bibliographystyle{splncs04}
\bibliography{egbib}

\section*{Supplementary Material} 
\setcounter{section}{0}
\section{Introduction}
In the supplementary material, we present more details about the experiments in our paper. 
The detailed average accuracy and harmonic accuracy values are reported in the tables. 
Following the CEC paper \cite{zhang2021few}, we also report on the performance dropping rate (PD).
The PD measures the absolute accuracy decrease between the base learning and the last incremental session.

\section{Detailed Results}
\begin{table*}[tpb]
	\centering
	\scriptsize{
		\caption{Experimental results for the \emph{8-step 5-way 5-shot} FSCIL protocol on the CIFAR100 dataset.
			The performance dropping rate (PD) measures the absolute accuracy decrease between the base learning and the last incremental session.
			The $*$ indicates results reported in \cite{zhang2021few,tao2020few} and the $\ddagger$ indicates results from our implementation using the official published code.
		}
	\vspace{+15pt}
		\begin{tabular}{l|@{\hskip 0.4cm}c@{\hskip 0.28cm}c@{\hskip 0.4cm}c@{\hskip 0.4cm}c@{\hskip 0.4cm}c@{\hskip 0.4cm}c@{\hskip 0.4cm}c@{\hskip 0.4cm}c@{\hskip 0.4cm}c@{\hskip 0.4cm}|c}
			\hline
			& 0 & 1 & 2 & 3 & 4 & 5 & 6 & 7 & 8 & PD $\downarrow$ \\
			\hline
			\multicolumn{11}{c}{class-wise average accuracy} \\
			\hline
			Ft-CNN$^*$ & 64.1 & 36.9 & 15.4 & 9.8 & 6.7 & 3.8 & 3.7 & 3.1 & 2.7 & 61.4 \\
			\hline
			iCaRL$^*$ \cite{rebuffi2017icarl} & 64.1 & 53.3 & 41.7 & 34.1 & 27.9 & 25.1 & 20.4 & 15.5 & 13.7 & 50.4 \\
			\hline
			EEIL$^*$ \cite{castro2018end} & 64.1 & 53.1 & 43.7 & 35.2 & 29.0 & 25.0 & 21.0 & 17.3 & 15.9 & 48.2 \\
			\hline
			NCM$^*$ \cite{hou2019learning} & 64.1 & 53.1 & 44.0 & 37.0 & 31.6 & 26.7 & 21.2 & 16.8 & 13.5 & 50.6 \\
			\hline
			TOPIC$^*$ \cite{tao2020few} & 64.1 & 55.9 & 47.1 & 45.2 & 40.1 & 36.4 & 34.0 & 31.6 & 29.4 & 34.7 \\
			\hline
			CEC$^*$ \cite{zhang2021few} & 73.1 & 68.9 & 65.3 & 61.2 & 58.1 & 55.6 & 53.2 & 51.3 & 49.1 & \textbf{24.0} \\
			\hline
			CEC$^\ddagger$ \cite{zhang2021few} & 74.0 & 68.1 & 64.2 & 60.6 & 57.3 & 54.8 & 52.5 & 50.3 & 48.1 & 25.9 \\
			\hline
			ALICE (Ours) & \textbf{79.0} & \textbf{70.5} & \textbf{67.1} & \textbf{63.4} & \textbf{61.2} & \textbf{59.2} & \textbf{58.1} & \textbf{56.3} & \textbf{54.1} & 24.9 \\
			\hline
			\multicolumn{11}{c}{harmonic accuracy} \\
			\hline
			CEC$^\ddagger$ \cite{zhang2021few} & - & 40.2 & 37.6 & 34.9 & 32.9 & 33.6 & 33.1 & 31.9 & 31.3 & - \\
			\hline
			ALICE (Ours) & - & \textbf{65.3} & \textbf{62.3} & \textbf{55.7} & \textbf{54.5} & \textbf{54.0} & \textbf{53.9} & \textbf{52.1} & \textbf{50.6} & - \\
			\hline
		\end{tabular}	
		\label{tab: cifar_5_shot}
	}
	\vspace{+15pt}
	\centering
	\scriptsize{
		\caption{Experimental results for the \emph{8-step 5-way 5-shot} FSCIL protocol on the miniImageNet dataset.
		}
		\begin{tabular}{l|@{\hskip 0.4cm}c@{\hskip 0.4cm}c@{\hskip 0.4cm}c@{\hskip 0.4cm}c@{\hskip 0.4cm}c@{\hskip 0.4cm}c@{\hskip 0.4cm}c@{\hskip 0.4cm}c@{\hskip 0.4cm}c@{\hskip 0.4cm}|c}
			\hline
			& 0 & 1 & 2 & 3 & 4 & 5 & 6 & 7 & 8 & PD $\downarrow$ \\
			\hline
			\multicolumn{11}{c}{class-wise average accuracy} \\
			\hline
			Ft-CNN$^*$ & 61.3 & 27.2 & 16.4 & 6.1 & 2.5 & 1.6 & 1.9 & 2.6 & 1.4 & 59.9 \\
			\hline
			iCaRL$^*$ \cite{rebuffi2017icarl} & 61.3 & 46.3 & 42.9 & 37.6 & 30.5 & 24.0 & 20.9 & 18.8 & 17.2 & 44.1 \\
			\hline
			EEIL$^*$ \cite{castro2018end}  & 61.3 & 46.6 & 44.0 & 37.3 & 33.1 & 27.1 & 24.1 & 21.6 & 19.6 & 41.7 \\
			\hline
			NCM$^*$ \cite{hou2019learning} & 61.3 & 47.8 & 39.3 & 31.9 & 25.7 & 21.4 & 18.7 & 17.2 & 14.2 & 47.1 \\
			\hline
			TOPIC$^*$ \cite{tao2020few} & 61.3 & 50.1 & 45.2 & 41.2 & 37.5 & 35.5 & 32.2 & 29.5 & 24.4 & 36.9 \\
			\hline
			CEC$^*$ \cite{zhang2021few} & 72.0 & 66.8 & 63.0 & 59.4 & 56.7 & 53.7 & 51.2 & 49.2 & 47.6 & 24.4 \\
			\hline
			CEC$^\ddagger$ \cite{zhang2021few} & 71.2 & 66.0 & 61.9 & 58.6 & 56.4 & 53.4 & 50.7 & 48.8 & 47.2 & \textbf{24.0} \\
			\hline
			ALICE (Ours) & \textbf{80.6} & \textbf{70.6} & \textbf{67.4} & \textbf{64.5} & \textbf{62.5} & \textbf{60.0} & \textbf{57.8} & \textbf{56.8} & \textbf{55.7} & 24.9 \\
			\hline
			\multicolumn{11}{c}{harmonic accuracy} \\
			\hline
			CEC$^\ddagger$ \cite{zhang2021few} & - & 34.6 & 31.0 & 29.0 & 31.8 & 28.9 & 26.9 & 27.5 & 28.1 & - \\
			\hline
			ALICE (Ours) & - & \textbf{64.9} & \textbf{58.9} & \textbf{56.4} & \textbf{55.4} & \textbf{52.7} & \textbf{50.8} & \textbf{51.0} & \textbf{50.9} & - \\
			\hline
		\end{tabular}	
		\label{tab: mini_imagenet_5_shot}
	}
	\vspace{+15pt}
	\centering
	\scriptsize{
		\caption{Experimental results for the \emph{10-step 10-way 5-shot} FSCIL protocol on the CUB200 dataset.
		}
		\begin{tabular}{l|@{\hskip 0.2cm}c@{\hskip 0.2cm}c@{\hskip 0.2cm}c@{\hskip 0.2cm}c@{\hskip 0.2cm}c@{\hskip 0.2cm}c@{\hskip 0.2cm}c@{\hskip 0.2cm}c@{\hskip 0.2cm}c@{\hskip 0.2cm}c@{\hskip 0.2cm}c@{\hskip 0.2cm}|c}
			\hline
			& 0 & 1 & 2 & 3 & 4 & 5 & 6 & 7 & 8 & 9 & 10 & PD $\downarrow$ \\
			\hline
			\multicolumn{13}{c}{class-wise average accuracy} \\
			\hline
			Ft-CNN$^*$ & 68.7 & 43.7 & 25.1 & 17.7 & 18.1 & 17.0 & 15.1 & 10.6 & 8.9 & 8.9 & 8.5 & 60.2 \\
			\hline
			iCaRL$^*$ \cite{rebuffi2017icarl} & 68.7 & 52.7 & 48.6 & 44.2 & 36.6 & 29.5 & 27.8 & 26.3 & 24.0 & 23.9 & 21.2 & 47.5 \\
			\hline
			EEIL$^*$ \cite{castro2018end}  & 68.7 & 53.6 & 47.9 & 44.2 & 36.3 & 27.5 & 25.9 & 24.7 & 24.0 & 24.1 & 22.1 & 46.6 \\
			\hline
			NCM$^*$ \cite{hou2019learning} & 68.7 & 57.1 & 44.2 & 28.8 & 26.7 & 25.7 & 24.6 & 21.5 & 20.1 & 20.1 & 19.9 & 48.8 \\
			\hline
			TOPIC$^*$ \cite{tao2020few} & 68.7 & 62.5 & 54.8 & 50.0 & 45.3 & 41.4 & 38.4 & 35.4 & 32.2 & 28.3 & 26.3 & 42.4 \\
			\hline
			Cheraghian \etal \cite{cheraghian2021semantic} & 68.2 & 60.5 & 55.7 & 50.5 & 45.7 & 42.9 & 40.9 & 38.8 & 36.5 & 34.9 & 33.0 & 35.2 \\
			\hline
			Cheraghian \etal \cite{cheraghian2021synthesized} & 68.8 & 59.4 & 59.3 & 55.0 & 52.6 & 49.8 & 48.1 & 46.3 & 44.3 & 43.4 & 43.2 & 25.6 \\
			\hline
			CEC \cite{zhang2021few} & 75.9 & 71.9 & 68.5 & 63.5 & 62.4 & 58.3 & 57.7 & 55.8 & 54.8 & 53.5 & 52.3 & 23.6 \\
			\hline
			CEC$^\ddagger$ \cite{zhang2021few} & 75.0 & 71.3 & 67.3 &  63.5 & 61.5 & 58.3 & 56.3 & 54.5 & 52.2 & 51.9 & 50.7 & 24.3 \\
			\hline
			ALICE (Ours) & \textbf{77.4} & \textbf{72.7} & \textbf{70.6} & \textbf{67.2} & \textbf{65.9} & \textbf{63.4} & \textbf{62.9} & \textbf{61.9} & \textbf{60.5} & \textbf{60.6} & \textbf{60.1} & \textbf{17.3} \\
			\hline
			\multicolumn{13}{c}{harmonic accuracy} \\
			\hline
			CEC$^\ddagger$ \cite{zhang2021few} & - & 59.6 & 52.6 & 46.6 & 48.1 & 45.0 & 44.7 & 44.4 & 42.3 & 44.2 & 43.9 & - \\
			\hline
			ALICE (Ours) & - & \textbf{70.0} & \textbf{65.6} & \textbf{59.3} & \textbf{59.6} & \textbf{57.6} & \textbf{58.9} & \textbf{58.8} & \textbf{57.8} & \textbf{58.8} & \textbf{59.0} & - \\
			\hline
		\end{tabular}	
		\label{tab: cub200_5_shot}
	}
\end{table*}

\begin{table*}[tpb]
	\centering
	\scriptsize{
		\caption{Experimental results for the \emph{8-step 5-way 1-shot} FSCIL protocol on the CIFAR100 dataset.
		}
	\vspace{+15pt}
		\begin{tabular}{l|@{\hskip 0.4cm}c@{\hskip 0.4cm}c@{\hskip 0.4cm}c@{\hskip 0.4cm}c@{\hskip 0.4cm}c@{\hskip 0.4cm}c@{\hskip 0.4cm}c@{\hskip 0.4cm}c@{\hskip 0.4cm}c|c}
			\hline
			& 0 & 1 & 2 & 3 & 4 & 5 & 6 & 7 & 8 & PD $\downarrow$ \\
			\hline
			\multicolumn{11}{c}{class-wise average accuracy} \\
			\hline
			CEC$^\ddagger$ \cite{zhang2021few} & 74.0 & 67.3 & 62.6 & 59.0 & 55.3 & 52.1 & 49.5 & 47.0 & 44.8 &  \textbf{29.2} \\
			\hline
			ALICE (Ours) & \textbf{79.0} & \textbf{71.0} & \textbf{66.4} & \textbf{62.2} & \textbf{58.1} & \textbf{54.7} & \textbf{52.0} & \textbf{49.8} & \textbf{47.5} & 31.5 \\
			\hline
			\multicolumn{11}{c}{harmonic accuracy} \\
			\hline
			CEC$^\ddagger$ \cite{zhang2021few} & - & 12.1 & 11.4 & 14.3 & 13.4 & 13.1 & 13.7 & 13.3 & 13.0 & - \\
			\hline
			ALICE (Ours) & - & \textbf{35.7} & \textbf{33.9} & \textbf{33.0} & \textbf{29.2} & \textbf{28.2} & \textbf{27.6} & \textbf{27.3} & \textbf{26.5} & - \\
			\hline
		\end{tabular}	
		\label{tab: cifar_1_shot}
	}
	\vspace{+15pt}
	\centering
	\scriptsize{
		\caption{Experimental results for the \emph{8-step 5-way 1-shot} FSCIL protocol on the miniImageNet dataset.
		}
		\begin{tabular}{l|@{\hskip 0.4cm}c@{\hskip 0.4cm}c@{\hskip 0.4cm}c@{\hskip 0.4cm}c@{\hskip 0.4cm}c@{\hskip 0.4cm}c@{\hskip 0.4cm}c@{\hskip 0.4cm}c@{\hskip 0.4cm}c@{\hskip 0.4cm}|c}
			\hline
			& 0 & 1 & 2 & 3 & 4 & 5 & 6 & 7 & 8 & PD $\downarrow$ \\
			\hline
			\multicolumn{11}{c}{class-wise average accuracy} \\
			\hline
			CEC$^\ddagger$ \cite{zhang2021few} & 71.2 & 66.3 & 61.6 & 57.6 & 54.0 & 50.8 & 48.2 & 45.9 & 43.7 & \textbf{27.5} \\
			\hline
			ALICE (Ours) & \textbf{80.6} & \textbf{70.7} & \textbf{65.8} & \textbf{61.8} & \textbf{58.4} & \textbf{55.3} & \textbf{52.4} & \textbf{50.7} & \textbf{48.6} & 32.0 \\
			\hline
			\multicolumn{11}{c}{harmonic accuracy} \\
			\hline
			CEC$^\ddagger$ \cite{zhang2021few} & - & 9.0 & 7.9 & 7.3 & 6.7 & 6.0 & 6.4 & 7.2 & 7.8 & - \\
			\hline
			ALICE (Ours) & - & \textbf{35.2} & \textbf{25.2} & \textbf{24.5} & \textbf{26.0} & \textbf{25.3} & \textbf{23.9} & \textbf{26.8} & \textbf{27.1} & - \\
			\hline
		\end{tabular}	
		\label{tab: mini_imagenet_1_shot}
	}
	\vspace{+15pt}
	\centering
	\scriptsize{
		\caption{Experimental results for the \emph{10-step 10-way 1-shot} FSCIL protocol on the CUB200 dataset.
		}
		\begin{tabular}{l|@{\hskip 0.25cm}c@{\hskip 0.25cm}c@{\hskip 0.25cm}c@{\hskip 0.25cm}c@{\hskip 0.25cm}c@{\hskip 0.25cm}c@{\hskip 0.25cm}c@{\hskip 0.25cm}c@{\hskip 0.25cm}c@{\hskip 0.25cm}c@{\hskip 0.25cm}c@{\hskip 0.25cm}|c}
			\hline
			& 0 & 1 & 2 & 3 & 4 & 5 & 6 & 7 & 8 & 9 & 10 & PD $\downarrow$ \\
			\hline
			\multicolumn{13}{c}{class-wise average accuracy} \\
			\hline
			CEC$^\ddagger$ \cite{zhang2021few} & 75.0 & \textbf{69.5} & \textbf{64.9} & \textbf{59.9} & \textbf{57.0} & \textbf{53.1} & 50.7 & 48.0 & 46.0 & 44.8 & 43.0 & 32.0 \\
			\hline
			ALICE (Ours) & \textbf{77.4} & 66.7 & 62.7 & 58.6 & 55.3 & \textbf{53.1} & \textbf{50.9} & \textbf{49.3} & \textbf{47.1} & \textbf{46.9} & \textbf{45.7} & \textbf{31.7} \\
			\hline
			\multicolumn{13}{c}{harmonic accuracy} \\
			\hline
			CEC$^\ddagger$ \cite{zhang2021few} & - & 32.5 & 31.8 & 24.8 & 27.2 & 25.5 & 25.4 & 24.5 & 24.3 & 26.0 & 25.5 & - \\
			\hline
			ALICE (Ours) & - & \textbf{40.8} & \textbf{38.4} & \textbf{33.4} & \textbf{33.0} & \textbf{33.9} & \textbf{33.8} & \textbf{34.9} & \textbf{33.2} & \textbf{36.3} & \textbf{36.3} & - \\
			\hline
		\end{tabular}	
		\label{tab: cub200_1_shot}
	}
\end{table*}

\end{document}